\documentclass{article} %
\usepackage{iclr2024_conference,times}

\usepackage{amsmath,amsfonts,bm}

\def\eqref#1{equation~\ref{#1}}

\def\1{\bm{1}}

\DeclareMathAlphabet{\mathsfit}{\encodingdefault}{\sfdefault}{m}{sl}
\SetMathAlphabet{\mathsfit}{bold}{\encodingdefault}{\sfdefault}{bx}{n}

\usepackage[utf8]{inputenc} %
\usepackage[T1]{fontenc}    %
\usepackage{hyperref}       %
\usepackage{url}            %
\usepackage{booktabs}       %
\usepackage{amsfonts}       %
\usepackage{nicefrac}       %
\usepackage{microtype}      %
\usepackage{xcolor}         %

\usepackage[ruled,vlined,linesnumbered,commentsnumbered,norelsize]{algorithm2e}
\usepackage{enumitem}

\usepackage{graphicx}
\usepackage{subfigure}

\usepackage{amsmath}
\usepackage{amssymb}
\usepackage{mathtools}
\usepackage{amsthm}
\usepackage{enumitem}
\usepackage{bbding}
\usepackage{multicol}
\usepackage{multirow}
\usepackage{makecell}
\usepackage{comment}
\usepackage{wrapfig}
\usepackage{bbding}
\usepackage{stfloats}

\definecolor{MyDarkBlue}{rgb}{0,0.5,1}
\definecolor{MyDarkGreen}{rgb}{0.02,0.6,0.02}
\definecolor{MyDarkRed}{rgb}{0.8,0.02,0.02}
\definecolor{MyDarkOrange}{rgb}{0.40,0.2,0.02}
\definecolor{MyPurple}{RGB}{111,0,255}
\definecolor{MyRed}{rgb}{1.0,0.0,0.0}
\definecolor{MyGold}{rgb}{0.75,0.6,0.12}
\definecolor{MyDarkgray}{rgb}{0.66, 0.66, 0.66}

\usepackage{hyperref}
\usepackage{url}
\usepackage{CJKutf8}

\newcommand{\myparagraph}[1]{\vspace{-8pt}\paragraph{#1}}

\newcommand{\rebuttal}[1]{{{#1}}}

\newcommand{\fullname}{latent intuitive physics}

\newcommand{\renderer}{PhysNeRF}

\title{Latent Intuitive Physics: Learning to Transfer Hidden Physics from A 3D Video}

\author{
Xiangming~Zhu\thanks{Equal contribution.}
\quad
Huayu~Deng\footnotemark[1]
\quad
Haochen~Yuan\footnotemark[1]
\quad
Yunbo~Wang\thanks{Corresponding author: Yunbo~Wang. }
\quad
Xiaokang~Yang\\
MoE Key Lab of Artificial Intelligence, AI Institute, Shanghai Jiao Tong University\\
{\tt \{xmzhu76, deng\_hy99, yuanhaochen, yunbow, xkyang\}@sjtu.edu.cn}\\
\textcolor{magenta}{\url{https://sites.google.com/view/latent-intuitive-physics/}}
}

\iclrfinalcopy %
\begin{document}

\maketitle

\vspace{-6pt}
\begin{abstract}
    \vspace{-2pt}
    We introduce latent intuitive physics, a transfer learning framework for physics simulation that can infer hidden properties of fluids from a single 3D video and simulate the observed fluid in novel scenes. Our key insight is to use latent features drawn from a learnable prior distribution conditioned on the underlying particle states to capture the invisible and complex physical properties. To achieve this, we train a parametrized prior learner given visual observations to approximate the visual posterior of inverse graphics, and both the particle states and the visual posterior are obtained from a learned neural renderer. The converged prior learner is embedded in our probabilistic physics engine, allowing us to perform novel simulations on unseen geometries, boundaries, and dynamics without knowledge of the true physical parameters. We validate our model in three ways: (i) novel scene simulation with the learned visual-world physics, (ii) future prediction of the observed fluid dynamics, and (iii) supervised particle simulation. Our model demonstrates strong performance in all three tasks.
\end{abstract}

\vspace{-6pt}
\section{Introduction}
\vspace{-3pt}

Understanding the intricate dynamics of physical systems has been a fundamental pursuit of science and engineering. Recently, deep learning-based methods have shown considerable promise in simulating complex physical systems~\citep{battaglia2016interaction,mrowca2018flexible,schenck2018spnets,li2018learning,ummenhofer2020lagrangian,sanchez2020learning,shao2022transformer,Prantl2022Conserving, han2022learning, guan2022neurofluid, li2023pacnerf}. However, most previous works focus on physics simulation with given accurate physical properties, which requires strong domain knowledge or highly specialized devices. 
\textit{Let us consider a question: Can we predict physical systems with limited knowledge of its physical properties? } If not, is it possible to transfer hidden physics present in readily accessible visual observations into learning-based physics simulators~\citep{li2018learning,ummenhofer2020lagrangian, sanchez2020learning,Prantl2022Conserving}?

Inspired by human perception, researchers in the field of AI have proposed a series of \textit{intuitive physics} methods~\citep{mccloskey1983intuitive, battaglia2013simulation,ehrhardt2019unsupervised,xu2019densephysnet,li2020visual} to solve this problem. A typical approach is to build the so-called ``\textit{inverse graphics}'' models of raw visual observations, which involves training learning-based physical simulators by solving the inverse problem of rendering 3D scenes~\citep{guan2022neurofluid, li2023pacnerf}. However, NeuroFluid \citep{guan2022neurofluid} needs to finetune the deterministic transition model in response to every physics dynamics associated with new-coming physical properties. PAC-NeRF~\citep{li2023pacnerf} adopts heuristic (rather than learnable) simulators and explicitly infers physical properties (such as the viscosity of fluids) given visual observations. It requires an appropriate initial guess of the optimized properties and specifying the fluid type (\textit{e.g.}, Newtonian \textit{vs}. non-Newtonian fluids).

\begin{figure}[t]
    \centering
    \vspace{-15pt}
    \includegraphics[width=0.99\textwidth]{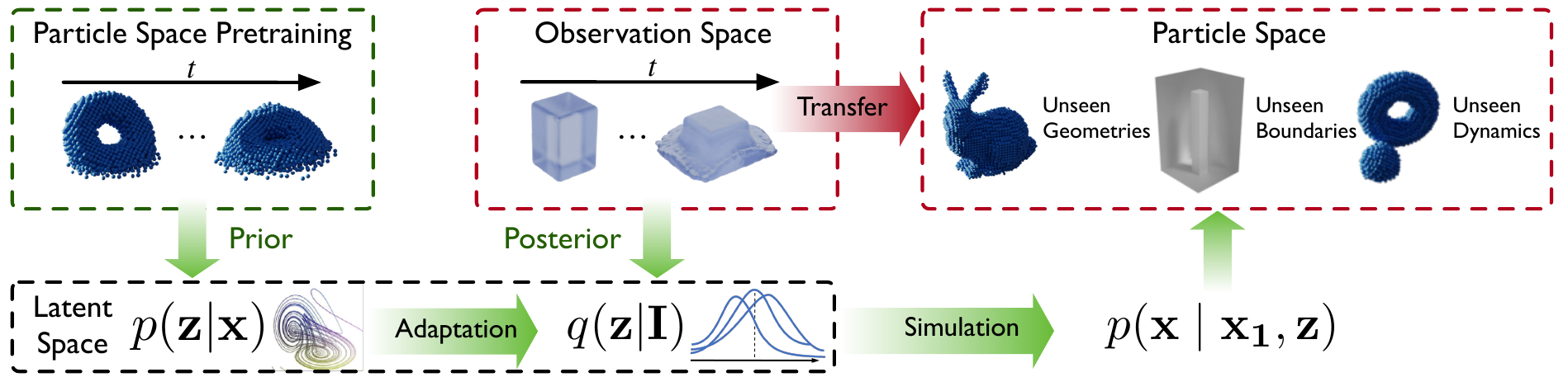}
    \vspace{-8pt}
    \caption{
    Our approach captures unobservable physical properties from image observations using a parametrized latent space and adapts them for simulating novel scenes with different fluid geometries, boundary conditions, and dynamics. To achieve this, we introduce a variational method that connects the particle space, observation space, and latent space for intuitive physical inference.
    }
    \label{fig:intro}
    \vspace{-10pt}
\end{figure}

\rebuttal{In this paper, we introduce the learning framework of \textit{\fullname}, which aims to infer hidden fluid dynamics from a 3D video, allowing for the simulation of the observed fluid in novel scenes without the need for its exact physical properties.}
The framework arises from our intuition that we humans can imagine how a fluid with a specific physical property will move given its initial state by watching a video showcase of it flowing, even though we do not explicitly estimate the exact values of physical properties.
The key idea is to represent the hidden physical properties in visual observations, which may be difficult to observe, using probabilistic latent states $z$ shown in Figure \ref{fig:intro}. 
The latent space connects the particle space and visual space to infer and transfer hidden physics with probabilistic modeling.
Specifically, our approach includes a probabilistic particle transition module $p(x^\prime | x,z)$\footnote{ Here, we use $x$ to indicate the historical states and $x^\prime$ to indicate the future states in the physical process.}, a physical prior learner, \rebuttal{a particle-based posterior estimator}, and a neural renderer, all integrated into a differentiable neural network.
The latent features are drawn from trainable marginal distributions $p(z|x)$ that are learned to approximate the visual posterior distribution $q(z|I)$ obtained from a learned neural renderer.
By employing probabilistic latent features, our model gains flexibility in modeling complex systems and is capable of capturing uncertainty in the data that a deterministic model may not be able to handle.
Once $p(z|x)$ is converged, we embed the prior learner in our probabilistic fluid simulator that is pretrained in particle space containing fluids with a wide range of physical properties. In this way, we transfer the hidden physics from visual observations to particle space to enable novel simulations of unseen fluid geometries, boundary conditions, and dynamics. In our experiments, we demonstrate the effectiveness of \textit{\fullname} by comparing it to strong baselines of fluid simulation approaches in novel scene simulation, future prediction of the observed dynamics, and supervised particle simulation tasks.

The contributions of this paper can be summarized as follows:
\begin{itemize}[leftmargin=*]
    \vspace{-6pt}
    \item \rebuttal{We introduce \textit{\fullname}, a learning-based approach for fluid simulation, which infers the hidden properties of fluids from 3D exemplars and transfers this knowledge to a fluid simulator.}
    \item We propose the first \textit{probabilistic particle-based fluid simulation network}, which outperforms prior works in particle-based simulation with varying physical properties.
\end{itemize}

\vspace{-3pt}
\section{Related Work}
\vspace{-3pt}

\paragraph{Learning-based simulation of particle dynamics.}

Recent research has introduced deep learning-based methods to accelerate the forward simulation of complex particle systems and address inverse problems. These methods have 
shown success in simulating rigid bodies~\citep{battaglia2016interaction,han2022learning}, fluids~\citep{belbute2020combining,sanchez2020learning,shao2022transformer,Prantl2022Conserving}, and deformable objects~\citep{mrowca2018flexible,li2018learning,sanchez2020learning,lin2022learning}. 
In the context of fluid simulation, DPI-Net~\citep{li2018learning} proposes dynamic graphs with multi-step spatial propagation, GNS~\citep{sanchez2020learning} uses message-passing networks, and TIE~\citep{shao2022transformer} uses a Transformer-based model to capture the spatiotemporal correlations within the particle system. 
Another line of work includes methods like CConv~\citep{ummenhofer2020lagrangian} and DMCF~\citep{Prantl2022Conserving}, which introduces higher-dimensional continuous convolution operators to model interactions between particles. 
Different from the approaches for other simulation scenarios for rigid and deformable objects, these models do not assume strong geometric priors such as object-centric transformations or pre-defined mesh topologies~\citep{pfaff2021learning, allen2022physical}.
However, these models are deterministic, assuming that all physical properties are measurable and focus on learning fluid dynamics under known physical properties.
Moreover, the stochastic components that commonly exist in the real physical world are not taken into account by the deterministic models.

\myparagraph{Intuitive physics learning with neural networks.}
Researchers have explored intuitive physics methods from a range of perspectives. These include heuristic models~\citep{gilden1994heuristic,sanborn2013reconciling}, probabilistic mental simulation models~\citep{hegarty2004mechanical,bates2015humans}, and the cognitive intuitive physics models~\citep{battaglia2013simulation,ullman2017mind}. 
Recent advances in deep learning typically investigate intuitive physics from different aspects. Some approaches adopt 3D videos to to downstream tasks, such as predicting multi-object dynamics~\citep{2022-driess-compNerf}, fluid dynamic~\citep{guan2022neurofluid, li2023pacnerf}, system identification~\citep{li2023pacnerf}, manipulation~\citep{simeonov2022neural, li20223d} or reasoning physics parameters~\citep{li2020visual, chen2021comphy,  le2023differentiable}. However, most works focus on rigid body dynamics~\citep{2022-driess-compNerf, li2020visual, le2023differentiable}. The most relevant work to our method is NeuroFluid~\citep{guan2022neurofluid} and PAC-NeRF~\citep{li2023pacnerf}, both focusing on fluid dynamics modeling with visual observation. 
NeuroFluid directly adopts the learning-based simulator from CConv~\citep{ummenhofer2020lagrangian}. Unlike our approach, it is deterministic and cannot handle the stochastic components or inaccessible physical properties in complex physical scenarios. 
PAC-NeRF employs specific non-learnable physics simulators tailored to different types of fluids. When using PAC-NeRF to solve inverse problems, users need to make an appropriate initialization for the optimized parameters based on the category of fluid observed.

\vspace{-4pt}
\section{Problem Formulation}
\vspace{-4pt}

\begin{figure*}[t]
\vspace{-15pt}
\begin{center}
\centerline{
\includegraphics[width=0.95\columnwidth]{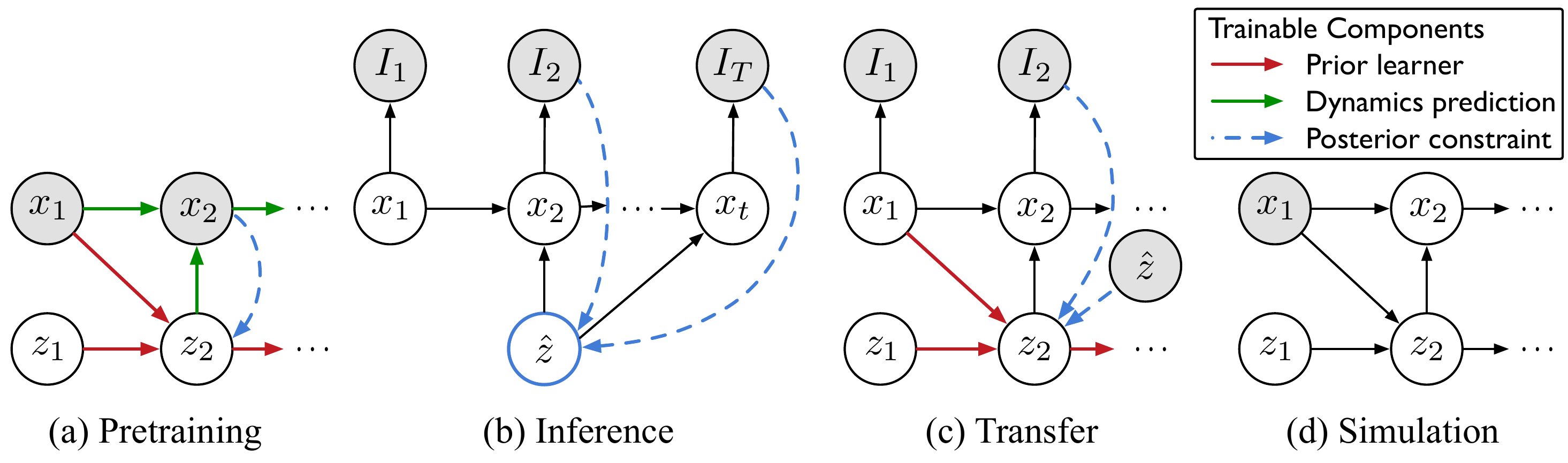}
}
\vspace{-10pt}
\caption{Graphical model of the \textit{pretraining--inference--transfer} pipeline of \fullname{}. (a) Particle-space pretraining for probabilistic fluid simulation. (b) Visual posterior optimization from visual observations with a photometric loss. (c) Adaptation of the prior learner to the converged visual posteriors $\hat z$. (d) Novel scene simulation with the adapted prior learner. The training parts are highlighted in color. \rebuttal{We present details of these training stages in Figure \ref{fig:pipeline_supp} in the appendix.}}
\label{fig:graphical}
\end{center}
\vspace{-25pt}
\end{figure*}

We study the inverse problem of fluid simulation, which refers to learning inaccessible physical properties by leveraging visual observations.
Specifically, we consider a dynamic system where we only have a single sequence of observations represented as multi-view videos $\{I_t^m\}_{t=1:T}^{m=1:M}$, where $I_t^m$ represents a visual observation received at time $t$ from view $m$. We want to predict the future states of the system when it appears in novel scenes.
Let $\mathbf{x}_t = (x_t^1, \ldots, x_t^N) \in \mathcal{X}$ be a state of the system at time $t$,
where $x_t^i=(p_t^i, v_t^i)$ represents the state of the $i^\text{th}$ particle that involves the position $p_t^i$ and velocity $v_t^i$, with $v_t^i$ being the time derivative of $p_t^i$. 
The dynamics of particles $\{\mathbf{x}_1,\ldots, \mathbf{x}_T\}$ is jointly governed by a set of physical properties, such as density, viscosity, and pressure. These properties are hidden and need to be inferred in visual observations.
To bridge the gap between particle simulators and the visual world with a varying set of physical properties in a unified framework, we introduce a set of latent features $\mathbf{z}_t= (z_t^1, \ldots, z_t^N)$, where $z_t^i$ is the latent feature attached to each particle. As shown in Figure~\ref{fig:graphical}(a), the particle state transition function can thus be represented as $\mathbf{x}_t \sim p(\mathbf{x}_{t-1}, \mathbf{z}_t)$, where $\mathbf{z}_t \sim p(\mathbf{x}_{1:t-1}, \mathbf{z}_{t-1})$.
As the explicit physical properties are inaccessible, we can infer latent distribution $p(\mathbf{z}_t \mid \mathbf{x}_{1:t})$ from $q(\mathbf{z} \mid I_{1:T})$. 
The final goal is to simulate novel scenes with new initial and boundary conditions based on learned 
physics (see Figure~\ref{fig:graphical}(d)).

\vspace{-4pt}
\section{Latent Intuitive Physics}
\vspace{-4pt}

\begin{figure*}[t]
\vspace{-15pt}
\begin{center}
\centerline{
\includegraphics[width=0.99\linewidth]{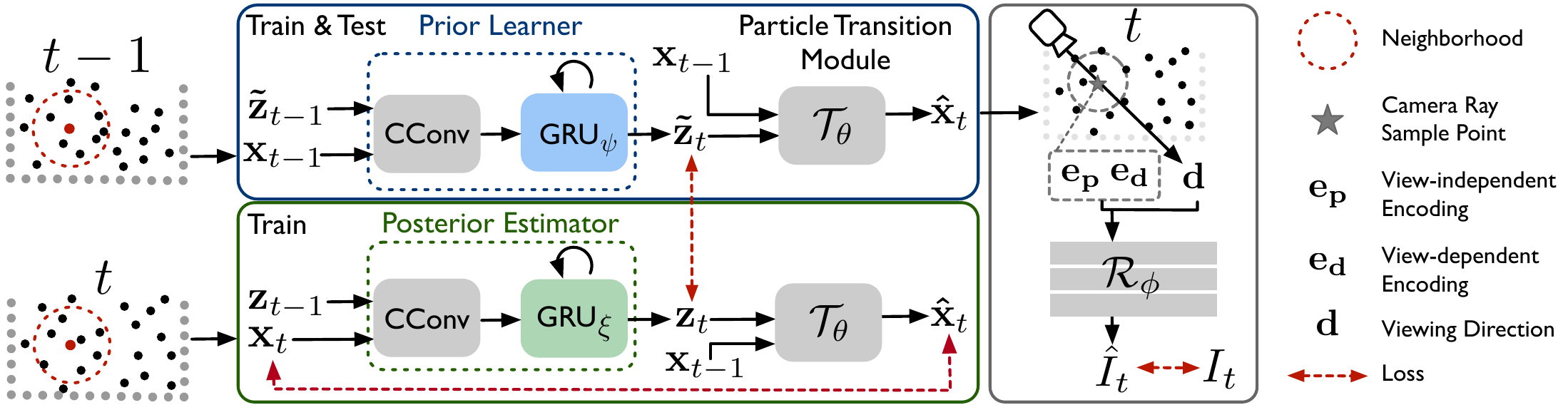}
}
\vspace{-7pt}
\rebuttal{\caption{Our model consists of four network components parametrized by $\theta,\psi,\xi,\phi$ respectively. We present the losses for pretraining the simulator and the renderer. For schematics of other training stages (\textit{i.e.}, visual posterior inference and prior adaptation), please refer to Figure~\ref{fig:pipeline_supp} in the appendix.
\label{fig:model_architecture}}
}
\end{center}
\vspace{-16pt}
\end{figure*}

In this section, we introduce \textit{latent intuitive physics} \rebuttal{for fluid simulation}, which enables the transfer of hidden fluid properties from visual observations to novel scenes.
As shown in Figure~\ref{fig:model_architecture}, our model consists of four network components: the probabilistic particle transition module ($\theta$), the physical prior learner ($\psi$), the particle-based posterior estimator ($\xi$), and the neural renderer ($\phi$). 
The training pipeline involves three stages shown in Figure~\ref{fig:graphical}---\textit{pretraining}, \textit{inference}, and \textit{transfer}:
\begin{enumerate}[leftmargin=*]
\renewcommand{\labelenumi}{\alph{enumi})}
    \vspace{-3pt}
    \item Pretrain the probabilistic fluid simulator on the particle dataset, which involves the particle transition module $p_\theta(\mathbf{x}_t \mid \mathbf{x}_{t-1}, \mathbf{z}_t)$, the prior learner $p_\psi(\tilde{\mathbf{z}}_t \mid \mathbf{x}_{1:t-1}, \tilde{\mathbf{z}}_{t-1})$, and the posterior $q_\xi(\mathbf{z}_t \mid \mathbf{x}_{1: t}, \mathbf{z}_{t-1})$. 
    The prior module reasons about latent features from historical particle states. 
    \vspace{-2pt}\item Infer the visual posterior latent features $\hat{\mathbf{z}}$ from consecutive image observations of a specific fluid, which is achieved by optimizing a differentiable neural renderer ($\phi$).
    \vspace{-2pt}\item Train the prior learner $p_\psi(\tilde{\mathbf{z}}_t \mid \mathbf{x}_{1:t-1}, \tilde{\mathbf{z}}_{t-1})$ to approximate the converged distribution of $\hat{\mathbf{z}}$, which enables the transfer of inaccessible physical properties from the visual world to the simulator.
\end{enumerate}
\vspace{-5pt}

\vspace{-3pt}
\subsection{Stage A: Probabilistic Fluid Simulator Pretraining} 
\vspace{-3pt}

Building a probabilistic model has significant advantages for fluid simulation: First, it allows us to predict fluid dynamics without knowing true physical parameters. Instead, it relies on inferring latent features from consecutive inputs. This is valuable because many complex physical phenomena naturally involve stochastic components.  Second, when provided with visual observations, we can seamlessly integrate the probabilistic fluid simulation into our variational inference method via the latent space. Next, we introduce how to infer physical properties from particle data.

The architecture of our probabilistic fluid simulator is shown in Figure~\ref{fig:model_architecture} (Left).
Particle states $\mathbf{x}_t$ itself is not a feature but simply defines the particle’s state in spatial space. We use the continuous convolution (CConv)~\citep{ummenhofer2020lagrangian} as a feature encoder to get feature representations of each particle. Inspired by traditional Smoothed Particle Hydrodynamics (SPH) methods, CConv predicts particle features by aggregating its neighbors' features in a continuous and smooth manner. Static particles such as boundaries, are processed similarly but with the particle positions and normal vectors as input (see the literature by \citet{ummenhofer2020lagrangian}).
Since the invisible physical properties cannot be inferred from a single state, we use a GRU to gather historical information and infer the distribution of the prior latents. The prior learner is trained along with a separate posterior estimator $q_\xi(\mathbf{z}_t \mid \mathbf{x}_{1:t}, \mathbf{z}_{t-1})$ (not used at test time). 
The models can be written as
\begin{equation}
    \text{Prior:} \ \tilde{\mathbf{z}}_t \sim \text{GRU}_{\psi}(\text{CConv}(\tilde{\mathbf{z}}_{t-1}, \mathbf{x}_{t-1})); \quad
    \text{Posterior:} \ \mathbf{z}_t \sim \text{GRU}_{\xi}(\text{CConv}(\mathbf{z}_{t-1}, \mathbf{x}_t)),
\label{eq: prior_posterior}
\end{equation}
where $\mathbf{z}_{t=1}$ and $\tilde{\mathbf{z}}_{t=1}$ are zero-initialized.
The posterior estimator takes $\mathbf{x}_t$ as input, \textit{i.e.}, the target of the prediction. 
 The prior and posterior latents are sampled from distinct Gaussian distributions, with their parameters determined by predicted means and variances by two GRUs.
During training, we align their distributions through KL divergence. 
Notably, our approach assumes time-varying and particle-dependent fluid properties, which aligns with conventional SPH methods (please refer to the work by \citet{bender2015divergence}). This approach empirically achieves better performance than optimizing a global latent variable, as demonstrated in our experiments.

For the particle transition module $p_\theta(\mathbf{x}_t \mid \mathbf{x}_{t-1}, \mathbf{z}_t)$, we adopt another CConv with additional inputs of $\mathbf{z}_t$ drawn from the inferred latent distribution. This allows the module to incorporate the previous states $\mathbf{x}_{t-1}$ and corresponding latent features for future prediction. 
As the stochastic physical component has been captured by $\mathbf{z}_t$, we employ a deterministic architecture for the particle transition module: \rebuttal{$\hat{\mathbf{x}}_t\triangleq\{(\hat{p}_t^i, \hat{v}_t^i)\}_{i=1:N}  \sim \mathcal{T}_\theta(\mathbf{x}_{t-1}, \mathbf{z}_{t})$}.
During training, the latent posteriors $\mathbf{z}_t$ are used as inputs of $\mathcal{T}_\theta$. At test time, we use the latent priors $\tilde{\mathbf{z}}_{t}$ instead.
The fluid simulator is trained with 
\begin{equation}
\mathcal{L}_{\theta, {\psi}, {\xi}} =  \ \mathbb{E} \Big[
 \frac{1}{N}\sum_{i=1}^N w_i \rebuttal{\left \| \hat{p}_t^i - p_t^i \right \|_2^\gamma} 
 +\beta \ \mathcal{D}_{K L}\left(q_{\xi}\left(\mathbf{z}_t \mid \mathbf{x}_{1: t}, \mathbf{z}_{t-1}\right) \| \ p_{\psi}\left(\tilde{\mathbf{z}}_t \mid \mathbf{x}_{1: t-1} , \tilde{\mathbf{z}}_{t-1}\right)\right) \Big].
\label{eq: loss}
\end{equation}
Similar to the previous work~\citep{li2018learning, Prantl2022Conserving, ummenhofer2020lagrangian, sanchez2020learning}, we use the $\|\cdot\|_2^\gamma$ error between the predicted position and the ground-truth positions, and weight it by the neighbor count to form the reconstruction loss. Specifically, we use $w_i=\exp(-\frac{1}{c} \mathcal{N}(\hat{p}_t^i))$, where $\mathcal{N}(\hat{p}_t^i)$ denotes the number of neighbors for the predicted particle $i$ and $c$ is the average neighbor count.

\vspace{-3pt}
\subsection{Stage B: Visual Posterior Inference} 
\vspace{-3pt}
Here we introduce how to solve the inverse problem by inferring scene-specific visual posteriors, where the visual observation governs the physical properties. In this stage, the pretrained particle transition module $\mathcal{T}_\theta$ infers the input visual posterior, which facilitates the adaptation of the prior learner in the latent space in Stage C. To this end, the particle transition module is combined with a differentiable neural renderer $\mathcal{R}_\phi$  that provides gradients backpropagated from the photometric error over sequences in observation space. Note that only visual observations are available in the following.

\myparagraph{Neural renderer.}
To enable joint modeling of the state-to-state function of fluid dynamics and the state-to-graphics mapping function, we integrate the probabilistic particle transition module with the particle-driven neural renderer (\renderer{}) in NeuroFluid~\citep{guan2022neurofluid} in a differentiable framework.
\renderer{} uses view-independent particle encoding $\mathbf{e}_\mathbf{p}$ and view-dependent particle encoding $\mathbf{e}_\mathbf{d}$ to estimate the volume density $\sigma$ and the color $\mathbf{c}$ of each sampled point along each ray $\mathbf{r}(t) = \mathbf{o} + t\mathbf{d}$, such that $(\mathbf{c}, \sigma) = \mathcal{R}_\phi(\mathbf{e}_\mathbf{p}, \mathbf{e}_\mathbf{d}, \mathbf{d})$. In this way, it establishes correlations between the particle distribution and the neural radiance field.
Unlike the original \renderer{}, we exclude the position of the sampled point from the inputs to the rendering network, which enhances the relationships between the fluid particle encodings and the rendering results. 
The neural renderer $\mathcal{R}_\phi$ is pretrained on multiple visual scenes so that it can respond to various particle-based geometries.

\myparagraph{Initial states estimation.}
NeuroFluid assumes known initial particle states. However, when only visual observations are available, estimating the initial particle states $\mathbf{x}_{t=1}$ becomes necessary.
These estimated initial states are used to drive the neural renderer ($\mathcal{R}\phi$) for generating visual predictions at the first time step and also to initiate the particle transition module ($\mathcal{T}_\theta$) for simulating subsequent states.
We estimate the initial particle positions using the voxel-based neural rendering technique~\citep{liu2020neural, sun2022direct, muller2022instant} at the first time step. During training, we maintain an occupancy cache to represent empty \textit{vs.} nonempty space and randomly sample fluid particles within each voxel grid in the visual posterior inference stage (see Appendix~\ref{sec:app_init_est} for details). 
\label{para: initial_pos_est}

\myparagraph{Optimization.} 
We first finetune the neural renderer $\mathcal{R}_\phi$ on current visual observation with initial state estimation $\hat{\mathbf{x}}_{t=1}$. 
Then the parameters of $\mathcal{T}_\theta$ and $\mathcal{R}_\phi$ are frozen and we initialize a set of visual posterior latents $\hat{\mathbf{z}}$, such that $\hat{\mathbf{x}}_t  = \mathcal{T}_\theta(\hat{\mathbf{x}}_{t-1}, \hat{\mathbf{z}})$. 
In practice, we attach a particle-dependent Gaussian distribution $\mathcal{N}(\hat{\mu}^i, \hat{\sigma}^i)$ with trainable parameters to each particle $i$. 
At each time step, we sample $\hat{z}^i \sim \mathcal{N}(\hat{\mu}^i, \hat{\sigma}^i)$ to form $\hat{\mathbf{z}} = (\hat{z}^1, \ldots, \hat{z}^N)$. 
The output of the neural renderer $\mathcal{R}_\phi$ is denoted as $\hat{\mathbf{C}}(\mathbf{r}, t)$. 
We optimize the distributions of the visual posterior latent $\hat{\mathbf{z}}$ over the entire sequence by backpropagating the photometric error $\sum\nolimits_{\mathbf{r},t} \|\hat{\mathbf{C}}(\mathbf{r},t) - \mathbf{C}(\mathbf{r},t) \|$. 
We summarize the overall training algorithm in the Alg.~\ref{algo:algorithm} in the appendix.

\vspace{-3pt}
\subsection{Stage C: Physical Prior Adaptation} 
\label{sec: stage_c}
\vspace{-3pt}

The visual posteriors learned in the previous stage are specific to the estimated particles within the observed scene and cannot be directly applied to simulate novel scenes.
Therefore, in this stage, we aim to adapt the hidden physical properties encoded in the visual posterior to the physical prior learner $p_{\psi}$. 
Instead of finetuning the entire particle transition model as NeuroFluid~\citep{guan2022neurofluid} does, we only finetune the prior learner module.
\rebuttal{
Due to the unavailability of the ground truth supervision signal in particle space, 
tuning all parameters in the transition model in visual scenes might lead to overfitting problems, as the transition model may forget the pre-learned knowledge of feasible dynamics, or learn implausible particle transitions even if it can generate similar fluid geometries that are sufficient to minimize the image rendering loss.}
Specifically, we perform forward modeling on particle states $\hat{\mathbf{x}}_t$ by applying $\hat{\mathbf{x}}_t = \mathcal{T}_\theta(\mathbf{x}_{t-1}, \tilde{\mathbf{z}}_t)$, where $\tilde{\mathbf{z}}_t$ is sampled from the distribution $p_{\psi}(\tilde{\mathbf{z}}_t \mid \mathbf{x}_{1: t-1},\tilde{\mathbf{z}}_{t-1})$ predicted by the prior learner. 
To transfer the visual posterior to $p_\psi$, we finetune the prior learner by minimizing the distance between its generated distribution and the pre-learned visual posteriors $\{\mathcal{N}(\hat{\mu}^i, \hat{\sigma}^i)\}^{i=1:N}$ with $\mathcal{T}_\theta$ and $\mathcal{R}_\phi$ fixed. 
The volume rendering loss is still used for supervision as well.
The entire training objective is 
\begin{equation}
    \mathcal{L}_\psi = \sum\nolimits_{\mathbf{r},t} \|\hat{\mathbf{C}}(\mathbf{r}, t) - \mathbf{C}(\mathbf{r}, t) \| + \beta \ \mathcal{D}_{KL}\left({q}(\mathbf{\hat z}) \ \| \ p_{\psi}\left(\tilde{\mathbf{z}}_t \mid \mathbf{x}_{1: t-1} , \tilde{\mathbf{z}}_{t-1}\right)\right).
\label{eq: stage2}
\end{equation}
With the finetuned physical prior learner $p_{\psi}$, we can embed it into the probabilistic fluid simulator to perform novel simulations on unseen fluid geometries, boundary conditions, and dynamics with identical physical properties, which brings the simulation back to the particle space.

\vspace{-3pt}
\section{Experiments}
\vspace{-3pt}

\vspace{-3pt}
\subsection{Evaluation of Visual Physical Inference}
\label{sec:eval_visual}
\vspace{-3pt}

\paragraph{Settings. } 
To evaluate \textit{\fullname} for inferring and transferring unknown fluid dynamics from visual observations, we generate a \textit{single} sequence using \textit{Cuboid} as the fluid body (not seen during pretraining in Stage A). The fluid freely falls in a default container. The sequence contains $60$ time steps, where the first $50$ time steps are used for visual inference, while the last $10$ time steps are reserved for additional validation on future rollouts. We use Blender~\citep{blender} to generate multi-view images with $20$ randomly sampled viewpoints. 
We assess the performance of our fluid simulator after transferring hidden properties from visual scenes through:
(1) simulating novel scenes with fluid geometries that were not seen during training (\textit{Stanford Bunny}, \textit{Sphere}, \textit{Dam Break}),
(2) simulating novel scenes with fluid dynamics with previously unseen boundaries. We conduct experiments across three distinct sets of physical properties. In each evaluation set, we impose random rotation, scaling, and velocities on fluid bodies. 
Please refer to Appendix~\ref{sec:app_data} \&~\ref{sec:app_exp_visual} for more information on visual examples, novel scenes for evaluation, and implementation details.
Following \citet{ummenhofer2020lagrangian}, we compute the \rebuttal{average Euclidean distance from the ground truth particles to the closest predicted particles as the evaluation metric, where the average prediction error is $\bar{d}=\frac{1}{T \times N} \sum_t \sum_i \min _{\hat{p}_t} ||p_t^i - \hat{p}_t||_2$. Please see Appendix~\ref{sec:evaluation_metrics} for details.} 

\myparagraph{Compared methods.} 
We use four baseline models.
\textit{CConv}~\citep{ummenhofer2020lagrangian} learns particle dynamics without physical parameter inputs. 
\textit{NeuroFluid}~\citep{guan2022neurofluid} grounds fluid dynamics in observed scenes with a particle-driven neural render. 
\textit{PAC-NeRF}~\citep{li2023pacnerf} and \textit{System Identification (Sys-ID)} estimate explicitly physical parameters in the observed scenes. PAC-NeRF employs an MPM simulator. Sys-ID utilizes a CConv simulator that takes learnable physical parameters as inputs. It also employs the same neural renderer as our approach.
All models are trained on \textit{Cuboid} and tested on novel scenes. Please refer to Appendix~\ref{sec:visual_baselines} for more details.

\begin{table}[t]
\vspace{-15pt}
  \centering
  \caption{\rebuttal{Quantitative results of average prediction error $\bar{d}$ on unseen fluid geometries and boundary conditions given $\mathbf{x}_{t=1}$.}  \rebuttal{We present the mean and standard deviation of $10$ independent samples drawn from our model (for \textit{Ours} only).} \textit{Geometry} means unseen fluid initial positions while \textit{Boundary} means unseen boundary conditions. Each physical property set is trained with a single visual observation.}
  \vspace{5pt}
  \begin{small}   
  \setlength{\tabcolsep}{1.5mm}{}
  \begin{sc}
    \begin{tabular}{lcccccc}
    \toprule
     & \multicolumn{2}{c}{$\rho=2000, \nu=0.065$} & \multicolumn{2}{c}{$\rho=1000, \nu=0.08$} & \multicolumn{2}{c}{$\rho=500, \nu=0.2$} \\
\cmidrule{2-7}    Method & \multicolumn{1}{l}{Geometry} & \multicolumn{1}{l}{Boundary} & \multicolumn{1}{l}{Geometry} & \multicolumn{1}{l}{Boundary} & \multicolumn{1}{l}{Geometry} & \multicolumn{1}{l}{Boundary} \\
\cmidrule{1-7}     CConv  & \underline{52.49} & \underline{64.29} & \underline{51.40}  & \underline{56.33} & \underline{40.67} & 53.28 \\
    NeuroFluid & 65.01 & 73.55 & 59.79 & 60.46 & 40.88 & \underline{50.73} \\
    Sys-ID & 156.59 & 179.11 & 127.06 & 140.27 & 58.71 & 71.80 \\
    PAC-NeRF & 51.10 & 59.61 & 51.33 & 56.84 & 40.97 & 62.05 \\
    Ours  & \rebuttal{\textbf{34.54$\pm$0.55}} & \rebuttal{\textbf{39.86$\pm$0.90}} & \rebuttal{\textbf{33.11$\pm$0.50} }& \rebuttal{\textbf{37.79$\pm$0.84}} & \rebuttal{\textbf{39.03$\pm$0.79}} & \rebuttal{\textbf{47.25$\pm$1.26}} \\
    \bottomrule
    \end{tabular}%
    \end{sc}
    \end{small}
  \label{tab:unseen}%
  \vspace{-5pt}
\end{table}%

\begin{table}[t]
\vspace{-5pt}
  \centering
  \caption{\rebuttal{Quantitative results of the average future prediction error $\bar{d}$ on observed scenes.} \rebuttal{We compute the mean results and standard deviations of $10$ independent samples drawn from our model.} Notably, NeuroFluid learns to overfit the observed scene by tuning the entire particle transition simulator.}
  \vspace{5pt}
  \begin{small}
  \setlength{\tabcolsep}{2.8mm}{}
    \begin{sc}
    \begin{tabular}{lccc}
    \toprule
    Method & $\rho=2000,\nu =0.065$ &$ \rho=1000,\nu =0.08$ & $\rho=500,\nu =0.2$ \\
    \midrule
    CConv & 107.27 & 103.95 &  54.37\\
    NeuroFluid & 85.45 & 73.04 &  \textbf{33.22}\\
    Sys-ID & \underline{37.25} & \underline{36.28} & 42.71 \\
    PAC-NeRF & 44.42 & 42.82 &  50.05\\
    Ours  & \rebuttal{\textbf{32.41$\pm$0.17}} & \rebuttal{\textbf{32.97$\pm$0.71}} &  \rebuttal{\underline{41.15$\pm$0.71}}\\
    \bottomrule
    \end{tabular}
    \end{sc}
    \end{small}
  \label{tab:rollout}
\vspace{-5pt}
\end{table}

\begin{figure}[t] 
    \centering
    \vspace{-15pt}
    \includegraphics[width=\textwidth]{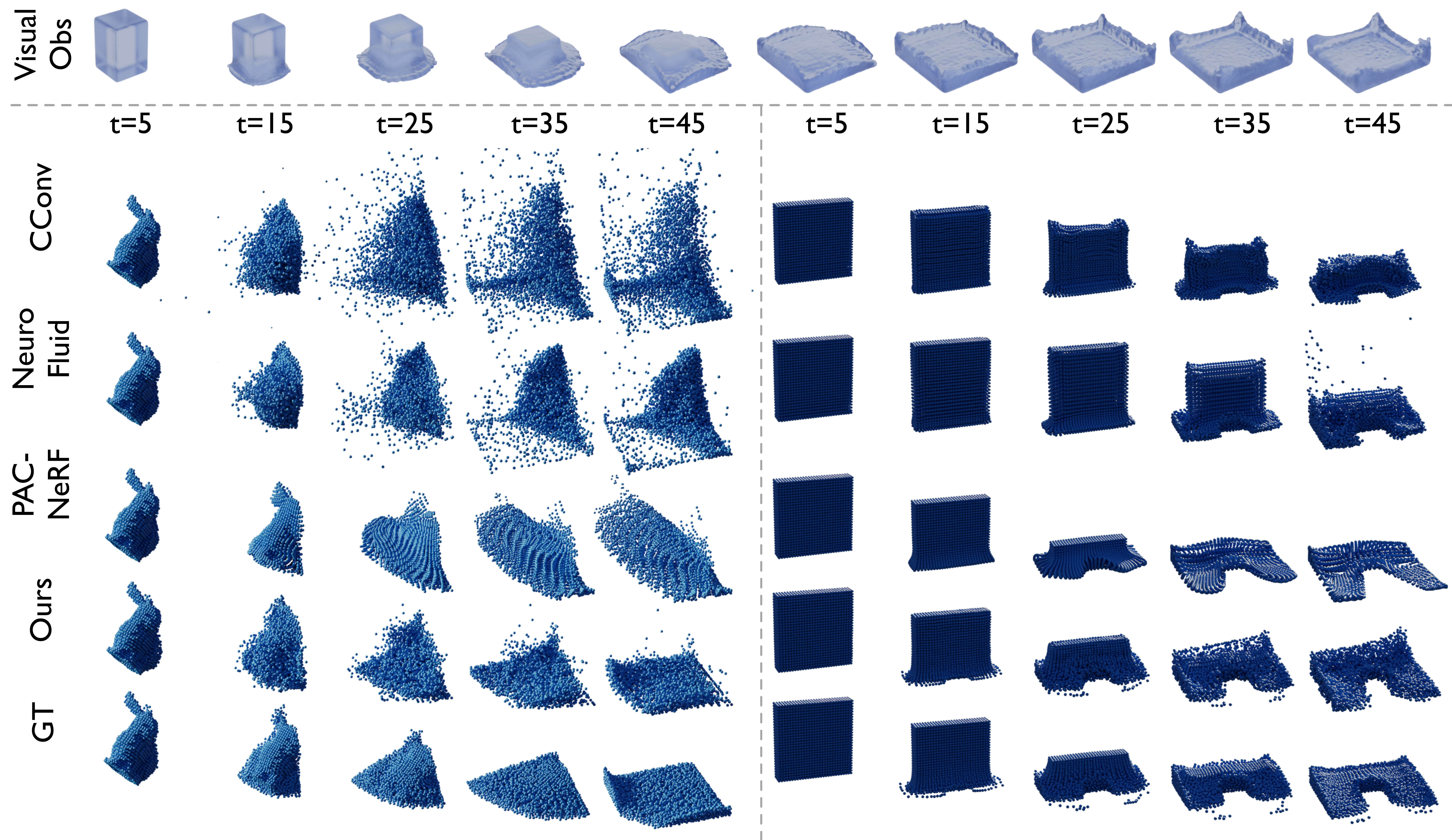}
    \vspace{-20pt}
    \caption{The first row shows the visual observation on the observed scene with physical parameters $\rho=2000, \nu=0.065$. Rows \rebuttal{2}-6 show qualitative results of simulated particles on novel scenes (Left: unseen geometries, Right: unseen boundaries). More qualitative results are provided in the appendix. 
    }
\label{fig:visualization}
\vspace{-10pt}
\end{figure}

\myparagraph{Novel scene simulation results.} 
We evaluate the simulation results of our approach given only the true initial particle states $\mathbf{x}_{t=1}$ of the novel scenes.
Table~\ref{tab:unseen} shows the average prediction error across all testing sequences.
\rebuttal{We predict each sequence $10$ times with different $\mathbf{z}_t$ drawn from the same prior learner and calculate the standard deviation of the errors.}
We can see that the adapted probabilistic fluid simulator from the visual posteriors significantly outperforms the baselines on novel scenes across all physical properties. 
Though our model is trained on scenes with the default fluid boundary, it shows the ability to generalize to unseen boundary conditions.
Figure~\ref{fig:visualization} showcases the qualitative results.
Among the baselines, Sys-ID underperforms in novel scene simulation, as it requires accurate parameter inference from visual observations.
In contrast, our model encodes the hidden properties with higher-dimensional latent variables, providing a stronger feature representation for the unknown properties in a physical system.
PAC-NeRF employs an MPM simulator, which inherently produces more accurate and stable simulation results than the learning-based simulators used by other models trained on limited data. Despite this advantage, PAC-NeRF tends to overfit the observed scenes and yields degraded performance when applied to novel scenes.
More results are shown in Appendix~\ref{sec:app_add_vis}.

\myparagraph{Future prediction results of the observed scenes.} We predict the particle dynamics of the observed scene \textit{Cuboid} for $10$ time steps into the future.
As shown in Table~\ref{tab:rollout}, our model performs best in most cases.
NeuroFluid slightly outperforms our model on $\rho=500, \nu=0.2$. Since NeuroFluid jointly optimizes the entire transition model and renderer on the observed scene, it is possible to overfit the observed scene and produce plausible future prediction results. However, it fails to generalize to novel scenes as shown in Table~\ref{tab:unseen}. \rebuttal{Unlike NeuroFluid, our approach adapts the physical prior learner to visual scenes, without training a probabilistic physics engine. By leveraging knowledge from the pretraining stage, the transition model is less prone to overfitting on the observed scene. This not only enhances the generalization ability but also significantly reduces the training burden.}

\vspace{-3pt}
\subsection{Evaluation of Probabilistic fluid simulator}
\label{sec:eval_simulator}
\vspace{-3pt}
To validate whether our approach can infer hidden physics from particle data in latent space, we evaluate the pretrained probabilistic fluid simulator ($\theta, \psi, \xi$) in a particle dataset generated with DFSPH~\citep{bender2015divergence}, which simulates fluids with various physical parameters (\textit{e.g.}, viscosity $\nu$, density $\rho$) falling in a cubic box. Each scene contains $273$-$19{,}682$ fluid particles and $200$ time steps. To assess the simulation performance under incomplete measurement of physical parameters, the true parameters are invisible to simulators. See the Appendix \ref{sec:app_particle_data} for more details.

We compare our probabilistic fluid simulator with four representative particle simulation approaches, based on graph neural networks, continuous convolution models, and Transformer, \textit{i.e.}, DPI-Net~\citep{li2018learning}, CConv~\citep{ummenhofer2020lagrangian}, DMCF~\citep{Prantl2022Conserving}, and TIE~\citep{shao2022transformer}. 
\rebuttal{Following~\citet{ummenhofer2020lagrangian}, given two consecutive input states $\mathbf{x}_{t-1:t}$, we compute the errors of the predicted particle positions w.r.t. the true particles: $d_{t+\tau}=\frac{1}{N} \sum_{i}  ||p_{t+\tau}^i - \hat{p}_{t+\tau}^i||_2$, for the next two steps ($\tau \in \{1,2\}$).}
To assess the long-term prediction ability, \rebuttal{we also calculate the average distance $\bar{d}$ from true particle positions to the predicted particles over the entire sequence.} The first $10$ states are given, and the models predict the following $190$ states. 
From Table~\ref{tab:particle}, our model performs best in both short-term and long-term prediction. 
The qualitative result of long-term prediction is shown in Figure~\ref {fig: longseq} in Appendix~\ref{sec:app_add_vis}. These results showcase that our probabilistic fluid simulation method provides an effective avenue for intuitive physics learning.

\begin{table}[t]
\vspace{-15pt}
  \centering
  \caption{
  Quantitative comparisons of fluid simulators on the particle dataset with inaccessible physical properties. 
  \rebuttal{We present the prediction errors for the next two frames $d_{t+1}$ and $d_{t+2}$, and the averaged prediction error $\bar{d}$ over the whole sequence for $190$ prediction time steps.} 
  }
  \vspace{5pt}
    \begin{small}   
    \setlength{\tabcolsep}{3mm}{}
    \begin{sc}
    \begin{tabular}{lccc}
    \toprule
    Methods & \rebuttal{$d_{t+1}$}   & \rebuttal{$d_{t+2}$}   & \rebuttal{$\bar{d}$}\\ %
    \midrule
    DPI-Net \citep{li2018learning} & 0.95  & 2.99  & 90.21 \\
    CConv \citep{ummenhofer2020lagrangian} & \underline{0.34}  & \underline{1.03}  & 44.79 \\
    DMCF \citep{Prantl2022Conserving}  & 0.54  & 1.23  & \underline{39.70} \\
    TIE  \citep{shao2022transformer} & 0.52  & 1.36  & 41.82 \\
    Ours  & \rebuttal{\textbf{0.31$\pm$0.003}}  & \rebuttal{\textbf{0.94$\pm$0.011}}  & \rebuttal{\textbf{38.37$\pm$0.860}} \\
    \bottomrule
    \end{tabular}%
    \end{sc}
    \end{small}   
  \label{tab:particle}
  \vspace{-10pt}
\end{table}

\begin{table}[t]
\centering
\caption{\rebuttal{Results of $\bar{d}$ on generalization to unseen fluid dynamics given $\mathbf{x}_{t=1}$.} \textit{Global Latent} is a variant of our model with a scene-specific global latent. }
  \vspace{5pt}
  \begin{small}
  \setlength{\tabcolsep}{2.5mm}{}
    \begin{sc}
    \begin{tabular}{lcccc}
        \toprule
        Methods & CConv & NeuroFluid & Global Latent & Ours \\
        \midrule
        Observed & 56.92 & \underline{54.67}
        & 60.63  & \rebuttal{\textbf{36.03 $\pm$ 0.80}} \\
        Unseen & \underline{46.83} & 54.50  & 90.51 & \rebuttal{\textbf{44.25 $\pm$ 1.36}} \\
        \bottomrule
    \end{tabular}%
    \end{sc}
    \end{small}
\label{tab:twodrops}%
\vspace{-10pt}
\end{table}

\begin{figure*}[!b] %
\vspace{-15pt}
\begin{center}
\centerline{
\includegraphics[width=\textwidth]{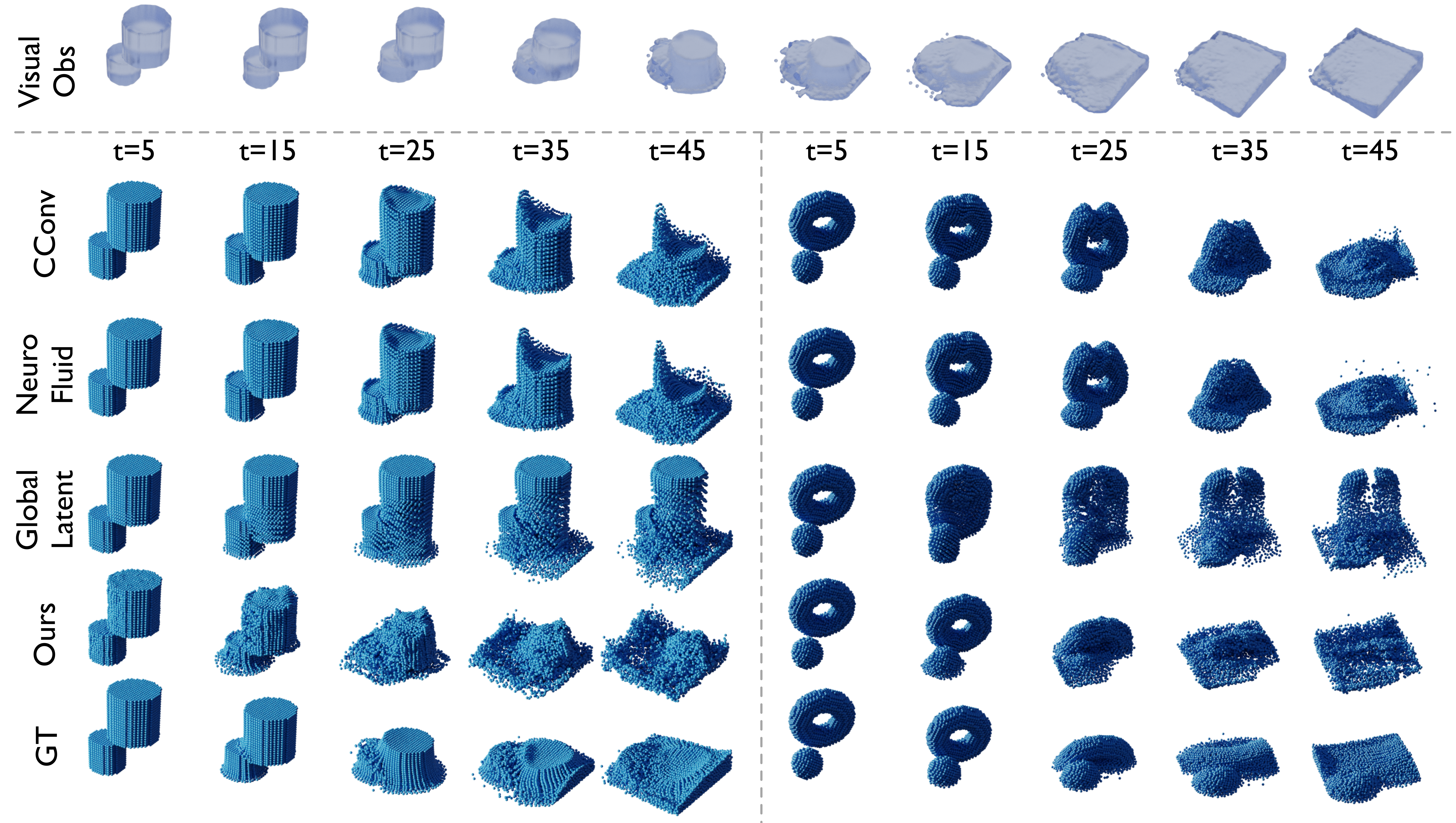}
}
\vspace{-10pt}
\caption{Qualitative results on generalization to unseen dynamics of heterogeneous fluids. We present simulation results on the observed scene (Left) and a novel scene (Right).}
\label{fig: Two drops}
\end{center}
\vspace{-20pt}
\end{figure*}

\vspace{-3pt}
\subsection{Generalization to Dynamics Discrepancies}
\label{sec:generalization}
\vspace{-3pt}

To validate the generalization ability of our approach across the discrepancies in fluid dynamics patterns, we consider a more complex scenario that contains a mixture of two different fluids. 
Specifically, we use the pretrained probabilistic fluid simulator discussed in Sec~\ref{sec:eval_simulator}, and adapt the model to a visual scene containing two heterogeneous fluids interacting with each other. 
We use two prior learners with the same initialization and a single particle transition module to learn the hidden physics of different fluid drops separately.
Table~\ref{tab:twodrops} and Figure~\ref{fig: Two drops} present both quantitative and qualitative results. Our approach showcases robust generalization ability when dealing with visual scenes containing fluid dynamics that diverge significantly from the patterns in the pretrained dataset.

Furthermore, we explore the performance of an alternative method that employs scene-specific, time-invariant latent variables (\textit{i.e.}, \textit{Global Latent} in Table~\ref{tab:twodrops} and Figure~\ref{fig: Two drops}). 
From these results, we find that optimizing time-varying latent features individually for each particle is more effective than optimizing two sets of global latents, each designated for a particular fluid type.

\vspace{-3pt}
\subsection{Ablation Study}
\label{sec:ablation}
\vspace{-3pt}

To verify the effectiveness of transferring the learned visual posterior $\hat{\mathbf{z}}$ to the physical prior learner, we experiment with different variants of our method. The result is shown in Table~\ref{tab:ablation}. \textit{w/o StageB} refers to the model without transferring posterior latent distribution $\hat{\mathbf{z}}$ to the physical prior learner and directly finetunes it on visual observations. \textit{w/o Stage C} refers to without adapting the physical prior learner. In this case, since we only have features $\mathbf{\hat z}$ attached to each particle, we cannot simulate novel scenes of various particle numbers.
The results show that the posterior latent distribution can make training more stable by restricting the range of distribution in latent space, and the transfer learning of the prior learner enables the prediction in novel scenes.   
In addition, we further investigate the performance gap between feeding the ground truth initial state and the estimated initial state to the probabilistic fluid simulator. We find that models utilizing estimated initial states yield comparable performance results when compared to those using ground truth initial states.

\begin{table}[t]
\vspace{-10pt}
  \centering
  \caption{Ablation study for each training stage in our pipeline. \rebuttal{We report mean prediction error ($\bar{d}$) resulting from the removal of Stage B, Stage C, or both, based on $10$ samples from our model. The future prediction error is reported on observed scenes and the prediction errors on unseen fluid geometries given $\mathbf{x}_{t=1}$.} \textbf{Ours$^\dagger$:} Using true initial states in the observed scenes for visual inference.}
    \vspace{5pt}
  \begin{small}
  \setlength{\tabcolsep}{2.5mm}{}
    \begin{sc}
    \begin{tabular}{lcccccc}
    \toprule
     & \multicolumn{2}{c}{$\rho = 2000,  \nu=0.065$} & \multicolumn{2}{c}{$\rho = 1000,  \nu=0.08$} & \multicolumn{2}{c}{$\rho = 500,  \nu=0.2$} \\
\cmidrule{2-7}    Methods & \multicolumn{1}{c}{Observed} & \multicolumn{1}{c}{Unseen} & \multicolumn{1}{c}{Observed} & \multicolumn{1}{c}{Unseen} & \multicolumn{1}{c}{Observed} & \multicolumn{1}{c}{Unseen} \\
    \midrule
    w/o Stage C & 33.22 & N/A     & \underline{32.72} & N/A     & \textbf{37.04} & N/A \\
    w/o Stage B & 37.55 & 42.43 & 35.75 & 46.98 & 41.71 & 41.40 \\
    Ours  & \underline{32.41}  & \underline{34.54} & 32.97 & \textbf{33.11} & 41.15 & \underline{39.03}\\
    Ours$^\dagger$ & \textbf{28.99} & \textbf{33.34} & \textbf{31.87} & \underline{38.84}  & \underline{39.70} & \textbf{38.41} \\
    \bottomrule
    \end{tabular}%
    \end{sc}
    \end{small}
  \label{tab:ablation}%
  \vspace{-10pt}
\end{table}%

\vspace{-3pt}
\section{Possibilities of Real-World Experiments}
\vspace{-3pt}

\begin{wrapfigure}{r}{6cm}%
\vspace{-12pt}
\centering
\includegraphics[width=0.4\textwidth]{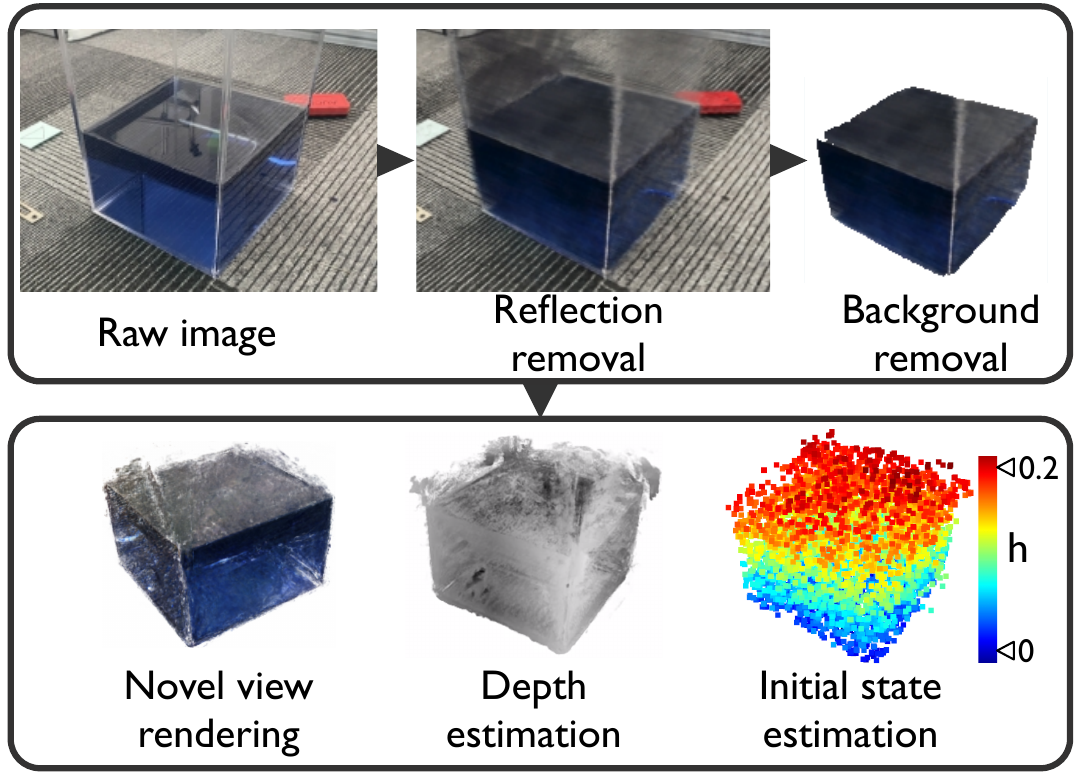}
\vspace{-8pt}
\caption{\rebuttal{Our pipeline and intermediate results for real-world experiments. We capture images of dyed water in a fluid tank and estimate the initial states. We will explore dynamic scenes in future work.}}
\vspace{-5pt}
\label{fig: real world main}
\end{wrapfigure}
The primary focus of this paper is to explore the feasibility of a new learning scheme for intuitive physics. Using synthetic data can greatly facilitate the evaluation of our model, as the simulation results can be directly quantified using particle states; whereas real-world scenes would necessitate more advanced fluid flow measurement techniques like \textit{particle image velocimetry}.
For a similar reason, earlier attempts like NeuroFluid and PAC-NeRF are also evaluated on synthetic data.
Nevertheless, we acknowledge that real-world validation is meaningful and challenging, and so make our best efforts to explore the possibility of implementing \textit{\fullname} in real-world scenarios. 
\rebuttal{As shown in Figure~\ref{fig: real world main}, we capture RGB images of dyed water in a fluid tank at a resolution of $1{,}200 \times 900$.
To cope with the complex and noisy visual environment, we adopt NeRFREN~\citep{Guo_2022_CVPR} to remove the reflection and refraction and segment the fluid body by SAM~\citep{kirillov2023segany}. The preprocessed images are then used to estimate fluid positions using our proposed initial state estimation module.
We carefully discuss the pipeline in Appendix \ref{sec:app_realworld} and include a video in the supplementary.}
Another notable challenge of the complete real-world experiments is to acquire high frame-rate images with synchronized cameras across multiple viewpoints. We leave this part for future work.

\vspace{-3pt}
\section{Conclusion}
\vspace{-3pt}

In this paper, we presented latent intuitive physics, which learns the hidden physical properties of fluids from a 3D video.
The key contributions include: 1) a probabilistic fluid simulator that considers the stochastic nature of complex physical processes, and 2) a variational inference learning method that can transfer the posteriors of the hidden parameters from visual observations to the fluid simulator.
Accordingly, we proposed the pretraining-inference-transfer optimization scheme for the model, which allows for easy transfer of visual-world fluid properties to novel scene simulation with various initial states and boundary conditions. 
We evaluated our model on synthetic datasets (similar to existing literature) and discussed its potential in real-world experiments.

\section*{Acknowledgments}

This work was supported by the National Natural Science Foundation of China (Grant No. 62250062, 62106144), the Shanghai Municipal Science and Technology Major Project (Grant No. 2021SHZDZX0102), the Fundamental Research Funds for the Central Universities, the Shanghai Sailing Program (Grant No. 21Z510202133), and the CCF-Tencent Rhino-Bird Open Research Fund.

\bibliography{iclr2024_conference}

\begin{thebibliography}{40}
\providecommand{\natexlab}[1]{#1}
\providecommand{\url}[1]{\texttt{#1}}
\expandafter\ifx\csname urlstyle\endcsname\relax
  \providecommand{\doi}[1]{doi: #1}\else
  \providecommand{\doi}{doi: \begingroup \urlstyle{rm}\Url}\fi

\bibitem[Allen et~al.(2022)Allen, Lopez-Guevara, Stachenfeld, Sanchez-Gonzalez, Battaglia, Hamrick, and Pfaff]{allen2022physical}
Kelsey~R Allen, Tatiana Lopez-Guevara, Kimberly Stachenfeld, Alvaro Sanchez-Gonzalez, Peter Battaglia, Jessica Hamrick, and Tobias Pfaff.
\newblock Physical design using differentiable learned simulators.
\newblock \emph{arXiv preprint arXiv:2202.00728}, 2022.

\bibitem[Bates et~al.(2015)Bates, Battaglia, Yildirim, and Tenenbaum]{bates2015humans}
Christopher Bates, Peter~W Battaglia, Ilker Yildirim, and Joshua~B Tenenbaum.
\newblock Humans predict liquid dynamics using probabilistic simulation.
\newblock In \emph{CogSci}, 2015.

\bibitem[Battaglia et~al.(2016)Battaglia, Pascanu, Lai, Jimenez~Rezende, et~al.]{battaglia2016interaction}
Peter Battaglia, Razvan Pascanu, Matthew Lai, Danilo Jimenez~Rezende, et~al.
\newblock Interaction networks for learning about objects, relations and physics.
\newblock In \emph{NeurIPS}, volume~29, 2016.

\bibitem[Battaglia et~al.(2013)Battaglia, Hamrick, and Tenenbaum]{battaglia2013simulation}
Peter~W Battaglia, Jessica~B Hamrick, and Joshua~B Tenenbaum.
\newblock Simulation as an engine of physical scene understanding.
\newblock \emph{Proceedings of the National Academy of Sciences}, 110\penalty0 (45):\penalty0 18327--18332, 2013.

\bibitem[Belbute-Peres et~al.(2020)Belbute-Peres, Economon, and Kolter]{belbute2020combining}
Filipe De~Avila Belbute-Peres, Thomas Economon, and Zico Kolter.
\newblock Combining differentiable pde solvers and graph neural networks for fluid flow prediction.
\newblock In \emph{ICML}, pp.\  2402--2411, 2020.

\bibitem[Bender \& Koschier(2015)Bender and Koschier]{bender2015divergence}
Jan Bender and Dan Koschier.
\newblock Divergence-free smoothed particle hydrodynamics.
\newblock In \emph{SCA}, pp.\  147--155, 2015.

\bibitem[Bender et~al.(2022)]{SPlisHSPlasH_Library}
Jan Bender et~al.
\newblock {SPlisHSPlasH Library}, 2022.
\newblock URL \url{https://github.com/InteractiveComputerGraphics/SPlisHSPlasH}.

\bibitem[Chen et~al.(2022)Chen, Yi, Li, Ding, Torralba, Tenenbaum, and Gan]{chen2021comphy}
Zhenfang Chen, Kexin Yi, Yunzhu Li, Mingyu Ding, Antonio Torralba, Joshua~B Tenenbaum, and Chuang Gan.
\newblock Comphy: Compositional physical reasoning of objects and events from videos.
\newblock In \emph{ICLR}, 2022.

\bibitem[Community(2018)]{blender}
Blender~Online Community.
\newblock \emph{Blender - a {3D} modelling and rendering package}.
\newblock Blender Foundation, Stichting Blender Foundation, Amsterdam, 2018.
\newblock URL \url{http://www.blender.org}.

\bibitem[Driess et~al.(2023)Driess, Huang, Li, Tedrake, and Toussaint]{2022-driess-compNerf}
Danny Driess, Zhiao Huang, Yunzhu Li, Russ Tedrake, and Marc Toussaint.
\newblock Learning multi-object dynamics with compositional neural radiance fields.
\newblock In \emph{CoRL}, volume 205, pp.\  1755--1768, 2023.

\bibitem[Ehrhardt et~al.(2019)Ehrhardt, Monszpart, Mitra, and Vedaldi]{ehrhardt2019unsupervised}
Sebastien Ehrhardt, Aron Monszpart, Niloy Mitra, and Andrea Vedaldi.
\newblock Unsupervised intuitive physics from visual observations.
\newblock In \emph{ACCV}, pp.\  700--716, 2019.

\bibitem[Gilden \& Proffitt(1994)Gilden and Proffitt]{gilden1994heuristic}
David~L Gilden and Dennis~R Proffitt.
\newblock Heuristic judgment of mass ratio in two-body collisions.
\newblock \emph{Perception \& Psychophysics}, 56:\penalty0 708--720, 1994.

\bibitem[Guan et~al.(2022)Guan, Deng, Wang, and Yang]{guan2022neurofluid}
Shanyan Guan, Huayu Deng, Yunbo Wang, and Xiaokang Yang.
\newblock Neurofluid: Fluid dynamics grounding with particle-driven neural radiance fields.
\newblock In \emph{ICML}, pp.\  7919--7929, 2022.

\bibitem[Guo et~al.(2022)Guo, Kang, Bao, He, and Zhang]{Guo_2022_CVPR}
Yuan-Chen Guo, Di~Kang, Linchao Bao, Yu~He, and Song-Hai Zhang.
\newblock Nerfren: Neural radiance fields with reflections.
\newblock In \emph{Proceedings of the IEEE/CVF Conference on Computer Vision and Pattern Recognition (CVPR)}, pp.\  18409--18418, June 2022.

\bibitem[Han et~al.(2022)Han, Huang, Ma, Li, Tenenbaum, and Gan]{han2022learning}
Jiaqi Han, Wenbing Huang, Hengbo Ma, Jiachen Li, Josh Tenenbaum, and Chuang Gan.
\newblock Learning physical dynamics with subequivariant graph neural networks.
\newblock In \emph{NeurIPS}, volume~35, pp.\  26256--26268, 2022.

\bibitem[Hegarty(2004)]{hegarty2004mechanical}
Mary Hegarty.
\newblock Mechanical reasoning by mental simulation.
\newblock \emph{Trends in cognitive sciences}, 8\penalty0 (6):\penalty0 280--285, 2004.

\bibitem[Kingma \& Ba(2015)Kingma and Ba]{kingma2014adam}
Diederik~P. Kingma and Jimmy Ba.
\newblock Adam: {A} method for stochastic optimization.
\newblock In \emph{ICLR}, 2015.

\bibitem[Kirillov et~al.(2023)Kirillov, Mintun, Ravi, Mao, Rolland, Gustafson, Xiao, Whitehead, Berg, Lo, Doll{\'a}r, and Girshick]{kirillov2023segany}
Alexander Kirillov, Eric Mintun, Nikhila Ravi, Hanzi Mao, Chloe Rolland, Laura Gustafson, Tete Xiao, Spencer Whitehead, Alexander~C. Berg, Wan-Yen Lo, Piotr Doll{\'a}r, and Ross Girshick.
\newblock Segment anything.
\newblock \emph{arXiv:2304.02643}, 2023.

\bibitem[Le~Cleac'h et~al.(2023)Le~Cleac'h, Yu, Guo, Howell, Gao, Wu, Manchester, and Schwager]{le2023differentiable}
Simon Le~Cleac'h, Hong-Xing Yu, Michelle Guo, Taylor Howell, Ruohan Gao, Jiajun Wu, Zachary Manchester, and Mac Schwager.
\newblock Differentiable physics simulation of dynamics-augmented neural objects.
\newblock \emph{IEEE Robotics and Automation Letters}, 8\penalty0 (5):\penalty0 2780--2787, 2023.

\bibitem[Li et~al.(2023)Li, Qiao, Chen, Jatavallabhula, Lin, Jiang, and Gan]{li2023pacnerf}
Xuan Li, Yi-Ling Qiao, Peter~Yichen Chen, Krishna~Murthy Jatavallabhula, Ming Lin, Chenfanfu Jiang, and Chuang Gan.
\newblock {PAC}-ne{RF}: Physics augmented continuum neural radiance fields for geometry-agnostic system identification.
\newblock In \emph{ICLR}, 2023.

\bibitem[Li et~al.(2019)Li, Wu, Tedrake, Tenenbaum, and Torralba]{li2018learning}
Yunzhu Li, Jiajun Wu, Russ Tedrake, Joshua~B Tenenbaum, and Antonio Torralba.
\newblock Learning particle dynamics for manipulating rigid bodies, deformable objects, and fluids.
\newblock In \emph{ICLR}, 2019.

\bibitem[Li et~al.(2020)Li, Lin, Yi, Bear, Yamins, Wu, Tenenbaum, and Torralba]{li2020visual}
Yunzhu Li, Toru Lin, Kexin Yi, Daniel Bear, Daniel~L.K. Yamins, Jiajun Wu, Joshua~B. Tenenbaum, and Antonio Torralba.
\newblock Visual grounding of learned physical models.
\newblock In \emph{ICML}, 2020.

\bibitem[Li et~al.(2022)Li, Li, Sitzmann, Agrawal, and Torralba]{li20223d}
Yunzhu Li, Shuang Li, Vincent Sitzmann, Pulkit Agrawal, and Antonio Torralba.
\newblock 3d neural scene representations for visuomotor control.
\newblock In \emph{CoRL}, pp.\  112--123, 2022.

\bibitem[Lin et~al.(2022)Lin, Wang, Huang, and Held]{lin2022learning}
Xingyu Lin, Yufei Wang, Zixuan Huang, and David Held.
\newblock Learning visible connectivity dynamics for cloth smoothing.
\newblock In \emph{CoRL}, pp.\  256--266, 2022.

\bibitem[Liu et~al.(2020)Liu, Gu, Zaw~Lin, Chua, and Theobalt]{liu2020neural}
Lingjie Liu, Jiatao Gu, Kyaw Zaw~Lin, Tat-Seng Chua, and Christian Theobalt.
\newblock Neural sparse voxel fields.
\newblock In \emph{NeurIPS}, volume~33, pp.\  15651--15663, 2020.

\bibitem[McCloskey et~al.(1983)McCloskey, Washburn, and Felch]{mccloskey1983intuitive}
Michael McCloskey, Allyson Washburn, and Linda Felch.
\newblock Intuitive physics: the straight-down belief and its origin.
\newblock \emph{Journal of Experimental Psychology: Learning, Memory, and Cognition}, 9\penalty0 (4):\penalty0 636, 1983.

\bibitem[Mildenhall et~al.(2020)Mildenhall, Srinivasan, Tancik, Barron, Ramamoorthi, and Ng]{mildenhall2021nerf}
Ben Mildenhall, Pratul~P. Srinivasan, Matthew Tancik, Jonathan~T. Barron, Ravi Ramamoorthi, and Ren Ng.
\newblock Nerf: Representing scenes as neural radiance fields for view synthesis.
\newblock In \emph{ECCV}, 2020.
\newblock URL \url{http://arxiv.org/abs/2003.08934v2}.

\bibitem[Mrowca et~al.(2018)Mrowca, Zhuang, Wang, Haber, Fei-Fei, Tenenbaum, and Yamins]{mrowca2018flexible}
Damian Mrowca, Chengxu Zhuang, Elias Wang, Nick Haber, Li~F Fei-Fei, Josh Tenenbaum, and Daniel~L Yamins.
\newblock Flexible neural representation for physics prediction.
\newblock In \emph{NeurIPS}, volume~31, 2018.

\bibitem[M{\"u}ller et~al.(2022)M{\"u}ller, Evans, Schied, and Keller]{muller2022instant}
Thomas M{\"u}ller, Alex Evans, Christoph Schied, and Alexander Keller.
\newblock Instant neural graphics primitives with a multiresolution hash encoding.
\newblock \emph{ACM Transactions on Graphics}, 41\penalty0 (4):\penalty0 1--15, 2022.

\bibitem[Pfaff et~al.(2021)Pfaff, Fortunato, Sanchez-Gonzalez, and Battaglia]{pfaff2021learning}
Tobias Pfaff, Meire Fortunato, Alvaro Sanchez-Gonzalez, and Peter Battaglia.
\newblock Learning mesh-based simulation with graph networks.
\newblock In \emph{International Conference on Learning Representations}, 2021.
\newblock URL \url{https://openreview.net/forum?id=roNqYL0_XP}.

\bibitem[Prantl et~al.(2022)Prantl, Ummenhofer, Koltun, and Thuerey]{Prantl2022Conserving}
Lukas Prantl, Benjamin Ummenhofer, Vladlen Koltun, and Nils Thuerey.
\newblock Guaranteed conservation of momentum for learning particle-based fluid dynamics.
\newblock In \emph{NeurIPS}, 2022.

\bibitem[Sanborn et~al.(2013)Sanborn, Mansinghka, and Griffiths]{sanborn2013reconciling}
Adam~N Sanborn, Vikash~K Mansinghka, and Thomas~L Griffiths.
\newblock Reconciling intuitive physics and newtonian mechanics for colliding objects.
\newblock \emph{Psychological review}, 120\penalty0 (2):\penalty0 411, 2013.

\bibitem[Sanchez-Gonzalez et~al.(2020)Sanchez-Gonzalez, Godwin, Pfaff, Ying, Leskovec, and Battaglia]{sanchez2020learning}
Alvaro Sanchez-Gonzalez, Jonathan Godwin, Tobias Pfaff, Rex Ying, Jure Leskovec, and Peter Battaglia.
\newblock Learning to simulate complex physics with graph networks.
\newblock In \emph{ICML}, pp.\  8459--8468, 2020.

\bibitem[Schenck \& Fox(2018)Schenck and Fox]{schenck2018spnets}
Connor Schenck and Dieter Fox.
\newblock Spnets: Differentiable fluid dynamics for deep neural networks.
\newblock In \emph{CoRL}, pp.\  317--335, 2018.

\bibitem[Shao et~al.(2022)Shao, Loy, and Dai]{shao2022transformer}
Yidi Shao, Chen~Change Loy, and Bo~Dai.
\newblock Transformer with implicit edges for particle-based physics simulation.
\newblock In \emph{ECCV}, pp.\  549--564, 2022.

\bibitem[Simeonov et~al.(2022)Simeonov, Du, Tagliasacchi, Tenenbaum, Rodriguez, Agrawal, and Sitzmann]{simeonov2022neural}
Anthony Simeonov, Yilun Du, Andrea Tagliasacchi, Joshua~B Tenenbaum, Alberto Rodriguez, Pulkit Agrawal, and Vincent Sitzmann.
\newblock Neural descriptor fields: Se (3)-equivariant object representations for manipulation.
\newblock In \emph{ICRA}, pp.\  6394--6400, 2022.

\bibitem[Sun et~al.(2022)Sun, Sun, and Chen]{sun2022direct}
Cheng Sun, Min Sun, and Hwann-Tzong Chen.
\newblock Direct voxel grid optimization: Super-fast convergence for radiance fields reconstruction.
\newblock In \emph{CVPR}, pp.\  5459--5469, 2022.

\bibitem[Ullman et~al.(2017)Ullman, Spelke, Battaglia, and Tenenbaum]{ullman2017mind}
Tomer~D Ullman, Elizabeth Spelke, Peter Battaglia, and Joshua~B Tenenbaum.
\newblock Mind games: Game engines as an architecture for intuitive physics.
\newblock \emph{Trends in cognitive sciences}, 21\penalty0 (9):\penalty0 649--665, 2017.

\bibitem[Ummenhofer et~al.(2020)Ummenhofer, Prantl, Thuerey, and Koltun]{ummenhofer2020lagrangian}
Benjamin Ummenhofer, Lukas Prantl, Nils Thuerey, and Vladlen Koltun.
\newblock Lagrangian fluid simulation with continuous convolutions.
\newblock In \emph{ICLR}, 2020.

\bibitem[Xu et~al.(2019)Xu, Wu, Zeng, Tenenbaum, and Song]{xu2019densephysnet}
Zhenjia Xu, Jiajun Wu, Andy Zeng, Joshua~B Tenenbaum, and Shuran Song.
\newblock Dense{PhysNet}: Learning dense physical object representations via multi-step dynamic interactions.
\newblock In \emph{RSS}, 2019.

\end{thebibliography}
\bibliographystyle{iclr2024_conference}

\newpage
\appendix
\section{Model Details}

\subsection{Overall Framework}
\label{sec:app_overall_framework}

\begin{figure}[ht]
\begin{center}
\centerline{
\includegraphics[width=\linewidth]{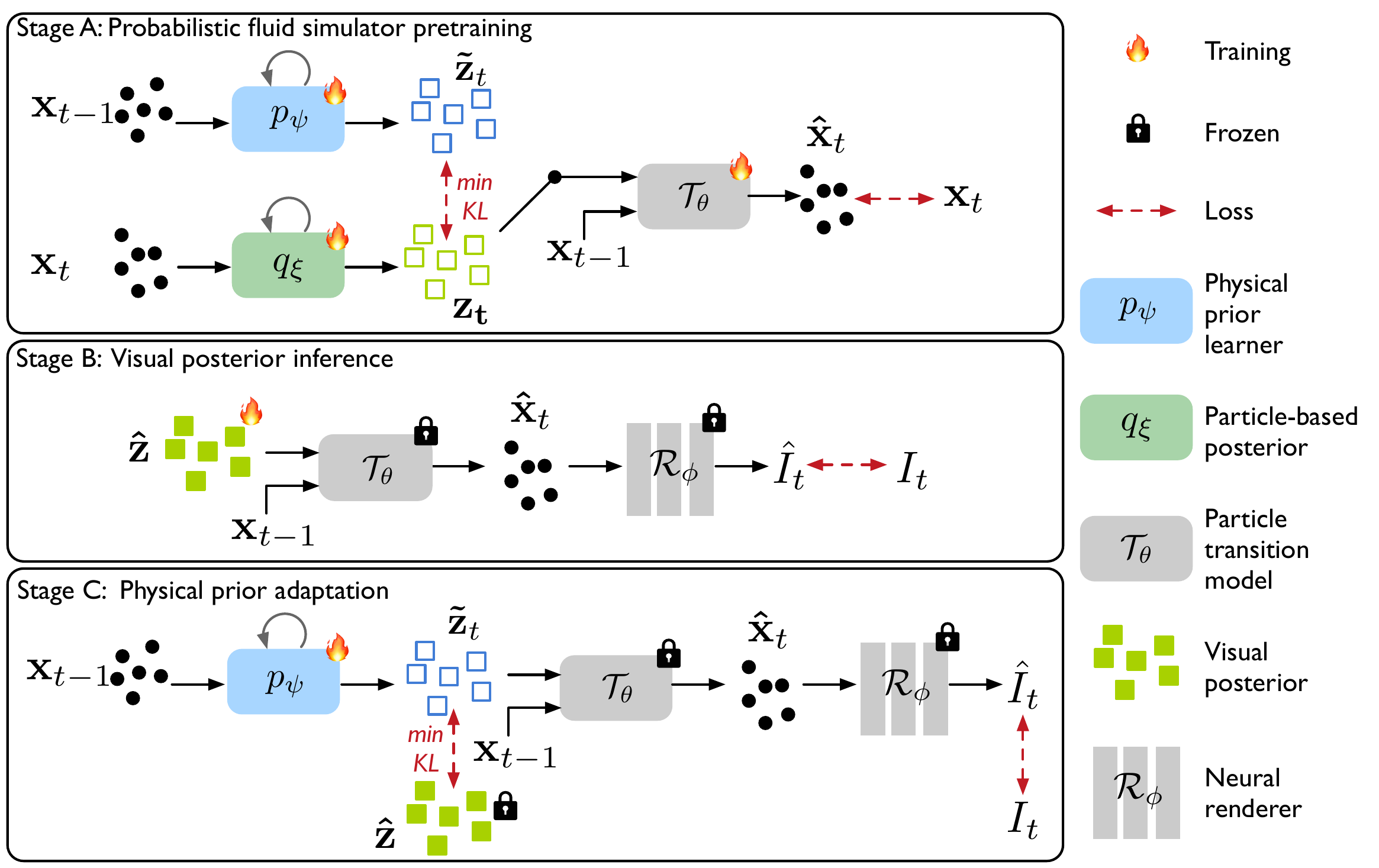}
}
\vspace{-10pt}
\rebuttal{
\caption{The overview of the proposed framework. The prior learner infers latent features from the data with the help of particle/visual posterior. The latent features are then used in the particle transition module to predict the next state. The neural renderer bridges the gap from the particle space to the visual observation space.}
\label{fig:pipeline_supp}
}
\end{center}
\end{figure}

Figure~\ref{fig:pipeline_supp} demonstrates the main architecture of the proposed framework\rebuttal{, and Table~\ref{tab:modules_supp} presents the summary of all model components, including formulation, input, output, training stage, and objective}. 
The entire model includes a probabilistic particle transition module ($\theta$), a physical prior learner ($\psi$), a particle-based posterior module ($\xi$), and a neural renderer ($\phi$). 
\rebuttal{
The physical prior learner $ p_\psi(\tilde{\mathbf{z}}_t   \mid   \mathbf{x}_{1: t-1}, \tilde{\mathbf{z}}_{t-1})$ infers latent features $\tilde{\mathbf{z}}_t$ from consecutive particle data, which the transition module $p_\theta(\mathbf{x}_t  \mid \mathbf{x}_{t-1}, \mathbf z_t)$ can use to predict the next state $\mathbf{x}_t$.
In the pre-training stage, a particle-based posterior $q_\xi(\mathbf{z}_t   \mid   \mathbf{x}_{1: t}, \mathbf{z}_{t-1})$  generates posterior $\mathbf{z}_t$ with the prediction target $ \mathbf{x}_t$ as input. It guides the prior learner to generate meaningful latent features that capture hidden properties essential for accurate predictions.
During visual inference and transfer in Stage B and C, the neural renderer $\mathcal{R}_\phi$ generates images according to particle positions and viewing directions, which enables photometric error to be backpropagated through the whole differentiable neural network.
In the visual posterior inference stage (Stage B), the visual posteriors $\mathbf{\hat z}$ are optimized by backpropagating image rendering errors. They serve as adaptation targets for the prior learner, facilitating efficient adaptation to the specific observed fluid.
}

\begin{table}[ht]
\centering
\vspace{-15pt}
\rebuttal{\caption{The overview of model components in our method.}
\label{tab:modules_supp}}
\vspace{5pt}
\resizebox{\textwidth}{!}{
\begin{tabular}{lll}
\toprule
\multirow{5}{*}{\makecell[l]{Particle\\transition\\module}} & Formulation              & $p_\theta(\mathbf{x}_t   \mid   \mathbf{x}_{t-1}, \mathbf z_t)$                \\
& Input                    &  Particle state $\mathbf{x}_{t-1}$   and   latent   feature $\mathbf{z}_t$                                \\
& Output                   & Next particle state   $\hat{\mathbf{x}}_t$                                        \\
& Training   stage         & Stage A                                                                     \\
& Objective & Predict the next state according   to particle states and latent features \\ \midrule
\multirow{5}{*}{\makecell[l]{Physical\\prior\\learner}}     & Formulation              & $\tilde{\mathbf{z}}_{t} \sim p_\psi(\tilde{\mathbf{z}}_t   \mid   \mathbf{x}_{1: t-1}, \tilde{\mathbf{z}}_{t-1})$       \\
& Input                    & Particle states $\mathbf{x}_{1:t-1}$   and  latent feature  $\tilde{\mathbf{z}}_{t-1}$                               \\
& Output                   & Prior latent feature   $\tilde{\mathbf{z}}_t$                               \\
& Training   stage         & Stage A \& C                                                                \\
& Objective & Narrow the gap with the posterior distribution                            \\ \midrule
\multirow{5}{*}{\makecell[l]{Particle-\\ based\\posterior}}   & Formulation              & $\mathbf{z}_t \sim q_\xi(\mathbf{z}_t   \mid   \mathbf{x}_{1: t}, \mathbf{z}_{t-1})$          \\
& Input                    & Future particle state $\mathbf{x}_{t}$, and previous states and features $(\mathbf{x}_{1:t-1}, \mathbf{z}_{t-1})$                   \\
& Output                   & Posterior latent feature   $\mathbf{z}_t$                                   \\
& Training   stage         & Stage A                                                                     \\
& Objective & Facilitate prior learner training when particle states are available                                                                        \\ \midrule
\multirow{5}{*}{\makecell[l]{Visual\\posterior}}   & Formulation              & $\hat{\mathbf{z}} \sim q(\hat{\mathbf{z}} \mid I_{1:T})$          \\
& Input                    & N/A                 \\
& Output                   & Particle-dependent feature                                \\
& Training   stage         & Stage B                                                                     \\
& Objective & Provide adaptation targets for the prior learner in subsequent Stage C                                                                        \\ \midrule
\multirow{5}{*}{\makecell[l]{Neural\\renderer}}            & Formulation              & $\mathcal{R}_\phi(\mathbf{e}_\mathbf{p}, \mathbf{e}_\mathbf{d}, \mathbf{d})$                                                          \\
& Input                    & Encoding based on predicted particle positions and viewing directions                                                \\
& Output                   & Rendered image                                                              \\
& Training   stage         & Pretrained on multiple visual scenes                                                                          \\
& Objective & Backpropagate the photometric error                                                                         \\ \bottomrule
\end{tabular}}
\end{table}

\subsection{Probabilistic Fluid Simulator}
The probabilistic fluid simulator is used to infer hidden physics and simulate various kinds of fluid dynamics. Specifically, as shown in Figure~\ref{fig:model_architecture}, our probabilistic fluid simulator consists of three modules: probabilistic particle transition module ($\theta$), physical prior learner ($\psi$) and particle-based posterior module ($\xi$). 

We use continuous convolution (CConv)~\citep{ummenhofer2020lagrangian} as the shared feature encoder for the prior learner ($\psi$) and the particle-based posterior module ($\xi$) to extract features from particle states. The CConv module predicts particle features in a smooth and continuous way: $(f * g)({x})=\frac{1}{n(x)} \sum_{i \in \mathcal{N}(x, R)} a\left(x^i, {x}\right) f_i g\left(\Lambda\left({x}^i-{x}\right)\right)$, with $f$ the input feature function and $g$ as filter function. The input consists of particle positions and the corresponding feature. The normalization $\frac{1}{n({x})}$ can be turned on with the normalization parameter. The per neighbor value $a({x}^i, {x})$ is a window function to produce a smooth response of the convolution under varying particle neighborhoods. It determines the $i^\text{th}$ particle feature by aggregating particle features given its neighbors' features. Thereby, the features of the neighbors are weighted with a kernel function depending on their relative position. The kernel functions themselves are discretized via a regular grid with spherical mapping and contain the learnable parameters. 
\rebuttal{To cope with boundary interactions, we follow the implementation of ~\citep{ummenhofer2020lagrangian}. Specifically, the feature encoder is realized with two separate CConv layers: One is to encode the fluid particles in the neighborhood of each particle location; The other one is to handle the virtual boundary particles in the same neighborhood. The features extracted from these processes are then concatenated to form the inputs for subsequent layers.}
The following GRUs can summarize the historical information and provide the inferred distribution of latent features about the system.
In practice, we use the mean of distributions at the last time step as input of the encoder CConv to avoid excessively noisy inputs.

\begin{figure*}[t]
\begin{center}
\centerline{
\includegraphics[width=0.8\linewidth]{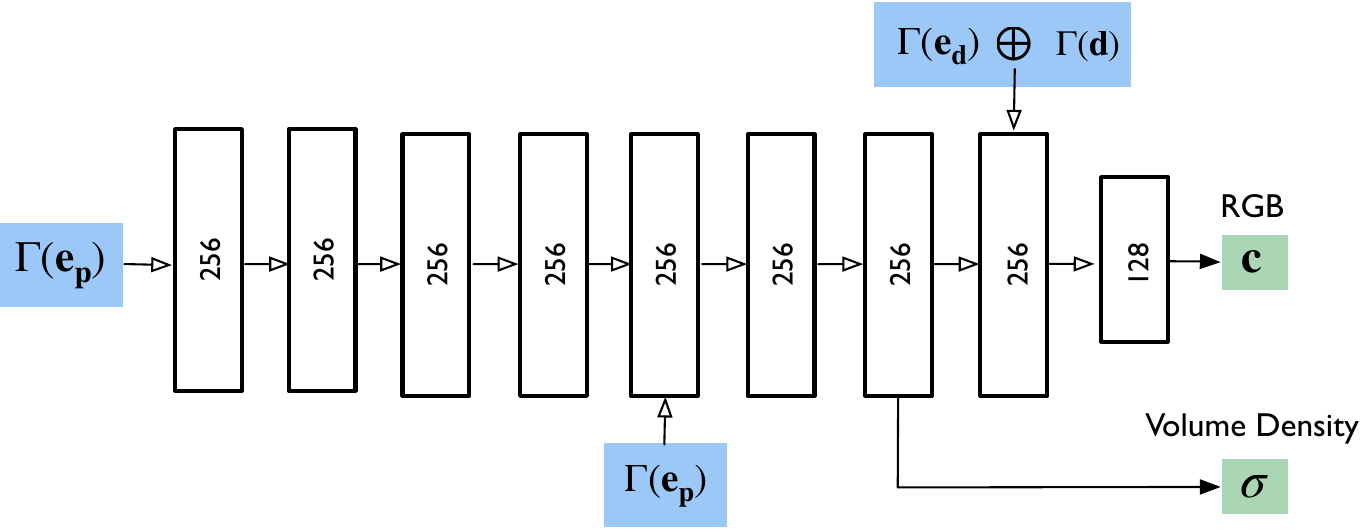}
}
\caption{Network architecture of \renderer{}. The inputs contain view-independent particle encoding $\mathbf{e}_\mathbf{p}$, view-dependent particle encoding $\mathbf{e}_\mathbf{d}$ and camera viewing direction $\mathbf{d}$. The outputs are volume density $\sigma$ and RGB color $\mathbf{c}$. }
\label{fig:renderer supp}
\end{center}
\vskip -0.1in
\end{figure*}

\subsection{Neural Renderer}

We present detailed model architecture of \renderer{}~\cite{guan2022neurofluid}. As shown in Figure~\ref{fig:renderer supp}, the network is based on fully-connected layers and similar to NeRF~\citep{mildenhall2021nerf}. \rebuttal{Unlike original NeRF, \renderer{} performs volume rendering according to the geometric distribution of neighboring physical particles along a given camera ray. 
For a sampled point in a ray, a ball query is conducted to identify neighboring fluid particles around the sampled point. 
These neighboring particles are then parameterized to obtain view-independent and view-dependent encodings $\mathbf{e}_\mathbf{p}$, $\mathbf{e}_\mathbf{d}$, which are used in the volume rendering subsequently. 
A multilayer perception (MLP) is trained to map these encodings along with view direction $\mathbf{d}$ to volume density $\sigma$ and emitted color $\mathbf{c}$ of each sampled point along each ray $\mathbf{r}(t) = \mathbf{o} + t\mathbf{d}$, such that $(\mathbf{c}, \sigma) = \mathcal{R}_\phi(\mathbf{e}_\mathbf{p}, \mathbf{e}_\mathbf{d}, \mathbf{d})$. Different from original \renderer{}, we exclude
the position of the sampled point from the inputs to the MLP network, which enhances the relationships between the fluid particle encodings and the rendering results.
The view-independent encodings $\mathbf{e}_\mathbf{p}$ includes 3 parts: fictitious particle center $\mathbf{p}_c$, soft particle density $\sigma_{\mathrm{p}}$ and radial deformation $\boldsymbol{v}_{\mathrm{D}}$, calculated as: }

\rebuttal{
\begin{equation}
    \begin{aligned}
        &\text{Fictitious particle center:} && {p}_\mathrm{c}=\frac{1}{K} \sum_{{p}^i \in \mathcal{N}(\mathbf{r}(t), r_s)} w_i {p}^i\\
&\text{Soft particle density:} && \sigma_{\mathrm{p}}=\sum_i w_i \\
&\text{Radial deformation:} && \boldsymbol{v}_{\mathrm{D}}=\frac{1}{K} \sum_i\left\|\left\|{p}^i - \mathbf{r}(t)\right\|-\frac{1}{K} \sum_i \left\|{p}^i - \mathbf{r}(t)\right\|\right\|_2  ,
    \end{aligned}
\end{equation}
}

\rebuttal{
where $\mathcal{N}(\mathbf{r}(t), r_s)$ is the ball query neighborhood within radius $r_s$ of the sampled camera ray point $\mathbf{r}(t)$ and  $w_i=\max (1-(\frac{\|{p}^i - \mathbf{r}(t)\|_2}{r_s})^3, 0)$. }
\rebuttal{
The view-dependent encoding $\mathbf{e}_\mathbf{d}$ includes the normalized view direction to the fictitious particle center, which is an important reference direction for the network to infer the refraction and reflection, calculated as: 
\begin{equation}
\boldsymbol{d}_\mathrm{c}=\left(p_\mathrm{c}-\mathbf{o}\right) /\left\|p_\mathrm{c}-\mathbf{o}\right\|_2.
\end{equation}
Finally, we take into account all the above physical quantities and derive the view-independent encoding and the view-dependent encoding as
\begin{equation}
    \mathbf{e}_\mathbf{p}=\left(p_\mathrm{c}, \sigma_{\mathrm{p}}, \boldsymbol{v}_{\mathrm{D}}\right), \quad \mathbf{e}_\mathbf{d}=\boldsymbol{d}_c.
\end{equation}
}

Following~\citep{mildenhall2021nerf, liu2020neural,sun2022direct}, we apply positional encoding $\Gamma(\cdot)$ to every input. In our experiments, we set the maximum encoded frequency $L = 10$ for $\Gamma(\mathbf{e}_{\mathbf{p}})$ and $L = 4$ for $\Gamma(\mathbf{e}_\mathbf{d}), \Gamma(\mathbf{d})$. We optimize \renderer{} in a coarse-to-fine manner~\citep{mildenhall2021nerf}. For the fine MLP network, the search radius of particle encoding of $\mathbf{e}_\mathbf{p}$ and $\mathbf{e}_\mathbf{d}$ is set as $3$ times the particle radius and we consider $20$ fluid particles within this search radius. For the coarse MLP network, 
the search radius scale and the number of encoding neighbors are set as $1.3$ times the parameters of the fine network.

\subsection{Hyperparameters}
Table~\ref{tab:hparams} shows the hyperparameters used in experiments.

\begin{table}[t]
\centering
\caption{Hyperparameters of the probabilistic fluid simulator and neural renderer}

    \begin{tabular}{lcc}
    \toprule
    \textbf{Name} & \textbf{Symbol} & \textbf{Value} \\
    \midrule
    \texttt{Stage A} \\
    \midrule
    KL loss scale & $\beta$ & 0.1 \\
    Distance $\|\cdot\|_2^{\gamma}$ & $\gamma$& 0.5 \\
    Gaussian latent dimension & --- & 8 \\
    Number of CConv blocks of shared feature encoder & --- & 3 \\
    Number of CConv blocks of Particle dynamics module & --- & 4 \\
    Number of GRU layers & --- & 1 \\
    Neighborhood radius & --- & 4.5 $\times$ particle radius \\
    \midrule
    \texttt{Stage B} \\
    \midrule
    Image size & H$\times $W & $400 \times 400$ \\
    Coarse sample points & $N_c$ & 64 \\
    Fine sample points & $N_f$ & 128 \\
    Particle encoding radius of $\mathbf{e}_{\mathbf{p}}$ and $\mathbf{e}_{\mathbf{d}}$  & $r_s$ & 3 $\times$ particle radius \\
    Particle encoding neighbors & $nn$ & 20 \\
    Coarse encoding scale & $s$ & 1.3 \\
    \midrule
    \texttt{Stage C} \\
    \midrule
    KL loss scale & $\beta$ & 0.01 \\
    \bottomrule
    \end{tabular}

\label{tab:hparams}
\end{table}

\section{Training Algorithm}
Algorithm~\ref{algo:algorithm} gives detailed descriptions of the computation flow of the training process.

\begin{algorithm}[t]
  \caption{Learning procedures in the visual scene}
  \label{algo:algorithm}
  \textbf{Given: }{Multi-view observation $\{I_t^{1:m}\}$, pretrained $\mathcal{T}_\theta$, prior learner $p_{\psi}$ (\texttt{Stage A}) and $\mathcal{R}_\phi$.} \\
  \textbf{Unknown: }{Particle state $\mathbf{x}_{t} = (x_t^1, \ldots, x_t^N)$ and scene-specific physical properties.} \\
\DontPrintSemicolon
  Estimate initial state $\hat{\mathbf{x}}_{t=1}=(\hat{x}_1^1, \ldots, \hat{x}_1^N)$ as described in Sec.~\ref{para: initial_pos_est}. \\
  Warmup $\mathcal{R}_\phi$ using rendering loss given $\hat{\mathbf{x}}_{t=1}$ on multi-view observation on $t=1$. \\
  Freeze $\mathcal{T}_\theta, \mathcal{R}_\phi$. \\
  \texttt{// Stage B: Optimize scene-specific latent features as visual posteriors} \\
  Initialize random Gaussian distribution $\mathcal{N}(\hat{\mu}^i, \hat{\sigma}^i)$ for each particle $i$. \\
  \While{not converged}{
    \For{time step $t = 1 \dots T$}{
        Sample particle feature for each particle: $\hat{z}^i_t \sim \mathcal{N}(\hat{\mu}^i, \hat{\sigma}^i)$ and produce $\hat{\mathbf{z}} = (\hat{z}^1, \ldots, \hat{z}^N)$.
        \\
        Simulate the next step $\hat{\mathbf{x}}_t = \mathcal{T}_\theta(\hat{\mathbf{x}}_{t-1}, \hat{\mathbf{z}})$ .\\
        Sample camera rays $\mathbf{r} \in {R}(\mathbf{P})$ and predict $\hat{\mathbf{C}}(\mathbf{r}, t)$ as described in Sec.~\ref{para: initial_pos_est}.\\
        Update the posterior distribution $\mathcal{N}(\hat{\mu}^i, \hat{\sigma}^i)$ for each particle $i$ using rendering loss $\mathcal{L}_t$ (Sec.~\ref{para: initial_pos_est}). \\
    }
  }
  \texttt{// Stage C: Finetune the pretrained physical prior learner $p_{\psi}$} \\
  Fix distribution $\mathcal{N}(\hat{\mu}^i, \hat{\sigma}^i)$ for each particle $i$. \\
  \While{not converged}{
    \For{time step $t = 1 \dots T$}{
        Sample particle feature for each particle: $\tilde{\mathbf{z}}_t \sim p_{\psi}(\tilde{\mathbf{z}}_t\mid \hat{\mathbf{x}}_{1:t-1}, \tilde{\mathbf{z}}_{t-1})$. \\ 
        Simulate the next step $\hat{\mathbf{x}}_t = \mathcal{T}_\theta(\hat{\mathbf{x}}_{t-1}, \tilde{\mathbf{z}}_t)$ .\\
        Predict $\hat{\mathbf{C}}(\mathbf{r}, t)$ by volume rendering as described in Sec.~\ref{para: initial_pos_est}.\\
        Update the prior learner $p_{\psi}$ using the loss function $\mathcal{L}_\psi$ in Eq.~\ref{eq: stage2}. \\
    }
  }
  Simulate novel scenes with the learned $\mathcal{T}_\theta$ and  $p_{\psi}$.
\end{algorithm}

\section{Datasets}
\label{sec:app_data}
\subsection{Particle Datasets}
\label{sec:app_particle_data}
This dataset is used for pretraining the probabilistic fluid simulator in Stage A. Following~\citep{ummenhofer2020lagrangian, Prantl2022Conserving}, we simulate the dataset with DFSPH~\citep{bender2015divergence} using the SPlisHSPlasH framework\footnote{https://github.com/InteractiveComputerGraphics/SPlisHSPlasH}. This simulator generates fluid flows with low-volume compression. The particle dataset contains 600 scenes, with 540 scenes used for training and 60 scenes reserved for the test set. In each scene, the simulator randomly places a fluid body of random shape in a cubic box (see \textit{Default Boundary} in Figure~\ref{fig:dataset}) and the fluid body freely falls under the influence of gravity and undergoes collisions with the boundary. The initial fluid body in each scene is applied with random rotation, scaling, and initial velocity. The simulator randomly samples physical properties viscosity $\nu$ and density $\rho$ for fluid bodies from uniform distribution $\mathcal{U}(0.01, 0.4)$ and $\mathcal{U}(500, 2000)$ respectively. The simulation lasts for 4 seconds, which consists of 200 time steps. In general, there are 273 $\sim$ 19682 fluid particles in each scene in this dataset.

\subsection{Generation of visual observations}
\label{sec:app_visual_data}
The visual observations are used in Stages B \& C.
Under each set of physical parameters, we generate a single 3D video of fluid dynamics with \textit{Cuboid} (unseen during Stage A) geometry. Each example contains $60$ time steps, where the most significant dynamic changes are included.
We use Blender~\citep{blender} to generate multi-view visual observations. Each fluid dynamic example is captured from $20$ randomly sampled viewpoints, with the cameras evenly spaced on the upper hemisphere containing the object. The first $50$ time steps are used for training, and the last $10$ time steps are used for the evaluation of future prediction. 
\subsection{Evaluation sets of visual physical inference}
\label{sec:app_visual_eval_data}
\begin{figure*}[t]
\begin{center}
\centerline{
\includegraphics[width=0.99\linewidth]{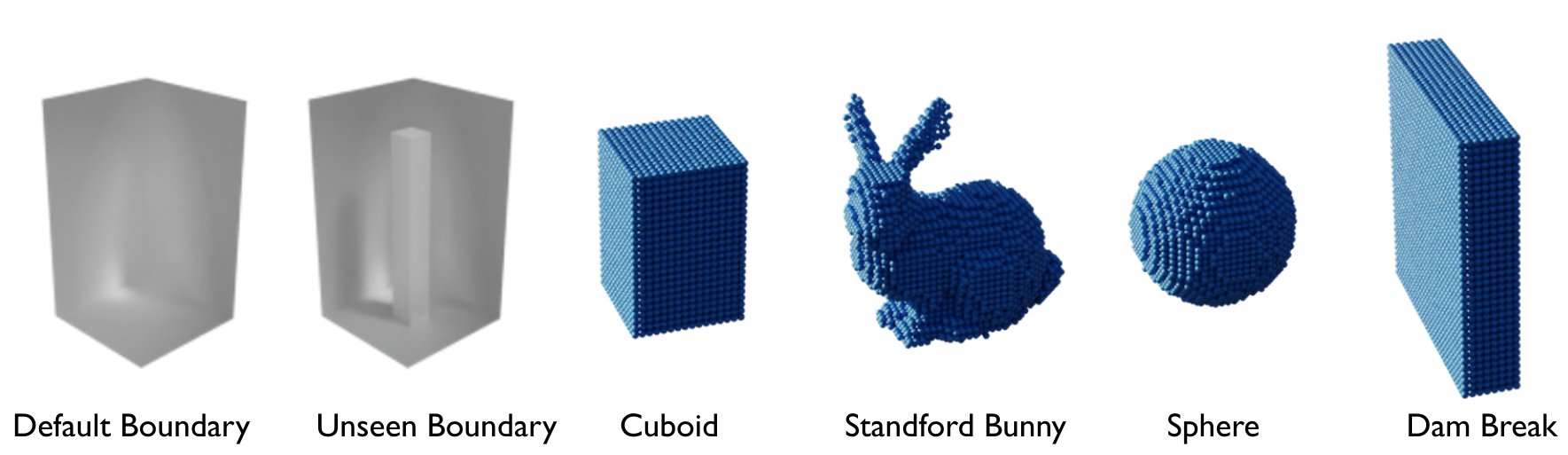}
}
\caption{Visualization of used boundaries and fluid bodies in visual physical inference. The leftmost container is the default boundary used in each scene in particle datasets and \textit{Cuboid}.}
\label{fig:dataset}
\end{center}
\vskip -0.1in
\end{figure*}

Our model and the baselines are evaluated on 3 challenging novel scene simulation tasks. Figure~\ref{fig:dataset} shows boundaries and fluid geometries used for evaluation.
\begin{itemize}[leftmargin=*]
\vspace{-5pt}
    \item \textbf{Unseen Fluid Geometries.} We use \textit{Standford Bunny}, \textit{Sphere}, \textit{Dam Break} that are unseen during pretraining as fluid bodies. The default boundary used in Stage A is used as a fluid container. We generate $50$ sequences of $60$ time steps under each set of physical parameters for evaluation. In each evaluation scene, random fluid geometry is chosen and random rotation, scaling, and velocities are imposed on the fluid body. 
    \item \textbf{Unseen Boundary conditions.} We use  \textit{Dam Break} fluid and unseen boundary for this task. This evaluation set features an out-of-domain boundary with a slender pillar positioned in the center. In each scene, the \textit{Dam Break} fluid collapses and strikes the pillar in the container, which was unseen during pretraining and from visual observations. Similarly, random rotation, scaling, and velocities are imposed on \textit{Dam Break} fluid, and 50 sequences of 60 time steps under each set of physical parameters for evaluation.
\end{itemize}
The horizontal initial velocity of the fluid body is randomly sampled from a uniform distribution of $\mathcal{U}(-2, 2)$, with a vertical initial velocity of $0$, and scale by a factor randomly sampled from another uniform distribution of $\mathcal{U}(0.8, 1.2)$. 
Table~\ref{tab:particle_number} shows the particle number of fluid bodies with no scaling applied. Table~\ref{tab:data_features} illustrates the features of the visual observation scene and evaluation sets.

\begin{table}[t]
  \centering
  \caption{The features of the visual observation scene and evaluation sets. \textit{Visual} means the visual observation scene, \textit{Geometry} means evaluation sets of unseen fluid geometries, and \textit{Boundary} means evaluation sets of unseen boundary conditions.}
  \begin{small} 
  \setlength{\tabcolsep}{1.mm}{}
    \begin{tabular}{lllccc}
    \toprule
          & Boundary & Fluid Body & Rotation, Scaling, Velocities & \#Sequence & Time Steps \\
    \midrule
    \multicolumn{1}{l}{\multirow{2}[1]{*}{Visual }} & \multirow{2}[1]{*}{Default} & \multicolumn{1}{l}{\multirow{2}[1]{*}{Cuboid}} & \multirow{2}[1]{*}{-} & \multirow{2}[1]{*}{1} & 60 \\
          &       &       &       &       & (50 for training )\\
    \multicolumn{1}{l}{\multirow{2}[0]{*}{Geometry}} & \multirow{2}[0]{*}{Default} & \multicolumn{1}{l}{Standford Bunny,} & \multirow{2}[0]{*}{\checkmark} & \multirow{2}[0]{*}{50} & \multirow{2}[0]{*}{60} \\
          &       & Sphere, Dam Break &       &       &  \\
    Boundary & Unseen & Dam Break & \checkmark     & 50    & 60 \\
    \bottomrule
    \end{tabular}
  \end{small}
  \label{tab:data_features}%
\end{table}%
\begin{table}[t]
    \caption{Particle number of different fluid bodies with no scaling applied.}
    \centering
    \begin{tabular}{lcccc}
    \toprule
        Dataset &  Cuboid & Stanford Bunny & Sphere & Dam Break \\
    \midrule
        Particle Numbers & 6144 & 8028 & 8211 & 10368 \\
    \bottomrule
    \end{tabular}
    \label{tab:particle_number}
\end{table}

\subsection{Dynamics Discrepancies}
In this task, we simulate two heterogeneous fluids with different physical properties. The generation of visual examples of this task is the same as previous tasks, but with two fluids of different physical properties interacting with each other. The evaluation set contains the same heterogeneous fluids but with unseen fluid geometries as fluid bodies. The evaluation set contains 50 sequences in total. Similarly, random rotation, scaling, and velocities are imposed on the fluid bodies in scenes in the evaluation set.

\section{Experimental Details}
\rebuttal{
\subsection{Evaluation Metrics}
\label{sec:evaluation_metrics}
We adopt the evaluation metric from the literature of CConv~\cite{ummenhofer2020lagrangian}. For all experiments but the short-term predictions in Table~\ref{tab:particle}, we compute the average Euclidean distance from the true particle positions ($p_t^i$) to the positions of the closest predicted particles ($\hat{p}_t$): }
\rebuttal{
$$d=\frac{1}{T \times N} \sum_t \sum_i \min _{\hat{p}_t} ||p_t^i - \hat{p}_t||_2$$ where $T$ is the prediction time horizon and $N$ is the number of particles.}
\rebuttal{
In particular, for the short-term prediction experiments ($n+1$ and $n+2$) in Table~\ref{tab:particle}, we compute an one-to-one mapping metric, the average error of the predicted particles w.r.t. the corresponding ground truth particles:}
\rebuttal{
$$d_t=\frac{1}{N} \sum_{i}  ||p_t^i - \hat{p}_t^i||_2$$ where $\hat{p}_t^i$ is the predicted position for particle $i$.
}
\subsection{Visual Physical Inference}
\label{sec:app_exp_visual}
\subsubsection{Baselines}
Our method is compared with the four baseline models below. For NeuroFluid and system identification, we adopt \renderer{} with the same architecture and hyperparameter in Table~\ref{tab:hparams}. All models are trained given the single 3D video of fluid dynamics with the \textit{Cuboid} geometry and evaluated on novel scenes given ground truth initial states. 
\label{sec:visual_baselines}
\begin{itemize}[leftmargin=*]
\vspace{-5pt}
    \item \textbf{CConv~\citep{ummenhofer2020lagrangian}:} Given the initial particle positions and velocities, CConv~\citep{ummenhofer2020lagrangian} uses a continuous convolution network that is performed in 3D space to simulate the particle transitions. However, the input to the CConv model only contains particle position and velocity such that it has a limitation that it can only perform simulation on fluid with identical physical parameters. We use this CConv model which has no additional input feature, pretrained on particle dataset in Sec.~\ref{sec:app_particle_data} and directly evaluate this model on novel scene simulations. 

    \item \textbf{NeuroFluid~\citep{guan2022neurofluid}:} A fully differentiable method for fluid dynamics grounding that links particle-based fluid simulation with particle-driven neural rendering in an end-to-end trainable framework. This approach links particle-based fluid simulation with particle-driven neural rendering in an end-to-end trainable framework, such that the two networks can be jointly optimized to obtain reasonable particle representations between them.  
    We adopt the CConv model pretrained on the particle dataset in Sec.~\ref{sec:app_particle_data} as the initial transition model and optimize NeuroFluid on \textit{Cuboid} scene in an end-to-end manner.
    
    \item \textbf{PAC-NeRF~\citep{li2023pacnerf}:} PAC-NeRF is a method that estimates physical parameters from multi-view videos with Eulerian-Lagrangian representation of neural radiance field and MPM simulator. We feed the physical parameters estimated on \textit{Cuboid} scene to the MPM simulator and use the given physical parameters to rollout on novel scenes.
    \item \textbf{System Identification (Sys-ID):}  For system identification, we train another CConv but with additional features of physical parameters for input including viscosity $\nu$ and density $\rho$. Sys-ID optimizes fluid properties by backpropagating the rendering loss through the trained transition network and PhysNeRF. The optimized physical parameters are used as model inputs to simulate novel scenes.
\end{itemize}
Note that baselines and our model are optimized over the entire visual sequence.
\subsubsection{Implementation Details}
We train our model and the baselines with multi-view observations on the fluid sequence of the \textit{Cuboid} geometry. 
Before Stage B, The \renderer{} is finetuned on visual observations for $100k$ steps with learning rate $3\mathrm{e}{-4}$ and exponential learning rate decay $\gamma=0.1$ given the estimated initial state and multi-view observation of the first frame. 
After that, we freeze $\mathcal{R}_\phi$ and $\mathcal{T}_\theta$ (pretrained on Particle Dataset) and infer the visual posterior by backpropagating the rendering loss. Then the physical prior learner $p_\psi$ is trained to adapt to the inferred visual posterior. The visual posterior latent and physical prior learner are separately optimized for $100k$ steps and $50k$ steps in Stage B and Stage C, with a learning rate of $1\mathrm{e}{-4}$ and a cosine annealing scheduler.

\subsubsection{Initial State Estimation}
\label{sec:app_init_est}
We estimate the initial state of fluid particles given multi-view observation on the first frame and then feed the estimated initial state to all simulators and the neural renderer. 
We estimate the initial particle positions using the voxel-based neural rendering technique~\citep{liu2020neural, sun2022direct, muller2022instant} and maintain an occupancy cache to represent empty \textit{vs.} nonempty space. At test time, we randomly sample fluid particles within each voxel grid to generate initial particle positions. 
To maintain spatial consistency between the estimated initial particle positions and the particle density generated by SPlisHSPlasH~\citep{SPlisHSPlasH_Library}, we employ the fluid particle discretization tools provided in SPlisHSPlasH to calculate the fluid particle count per unit volume. 
When estimating initial particle positions with new fluid scenes, we adopt the voxel-based neural renderer to predict the spatial volume occupied by the fluid, enabling the subsequent sampling of fluid particles according to the prescribed fluid particle density per unit volume. Since we do not apply random initial velocities to the fluid dynamics in visual examples, the initial velocities are set as zero.

\subsection{Evaluation of Probabilistic Fluid Simulator}

\subsubsection{Baselines on Particle Dataset}
\label{sec:particle_baselines}
We compare our probabilistic fluid simulator with four representative particle simulation approaches, based on GNN, continuous convolution, and Transformer, \textit{i.e.}, DPI-Net~\citep{li2018learning}, CConv~\citep{ummenhofer2020lagrangian}, DMCF~\citep{Prantl2022Conserving}, and TIE~\citep{shao2022transformer}. 
\begin{itemize}[leftmargin=*]
\vspace{-5pt}
    \item \textbf{DPI-Net~\citep{li2018learning}:} DPI-Net is a particle-based simulation method that combines multi-step spatial propagation, a hierarchical particle structure, and dynamic interaction graphs.
    \item \textbf{CConv~\citep{ummenhofer2020lagrangian}:} The method employs spatial convolutions as the primary differentiable operation to establish connections between particles and their neighbors. It predicts particle features and dynamics in a smooth and continuous way. %
    \item \textbf{DMCF~\citep{Prantl2022Conserving}:} DMCF imposes a hard constraint on momentum conservation by employing antisymmetrical continuous convolutional layers. It utilizes a hierarchical network architecture, a resampling mechanism that ensures temporal coherence.
    \item \textbf{TIE~\citep{shao2022transformer}: } A Transformer-based model that captures the complex semantics of particle interactions in an edge-free manner. The model adopts a decentralized approach to computation, wherein the processing of pairwise particle interactions is replaced with per-particle updates. The original method is conducted on the PyFlex dataset with less number of fluid particles ($\sim$ hundreds). However, on \textit{Particle Dataset} ($\sim$ thousands), the experiment on TIE leads to unacceptable memory cost. Therefore, we downsample the fluid particles in each scene of the \textit{Particle Dataset} to the ratio of $1/20$.
\vspace{-5pt}
\end{itemize}

\subsubsection{Implementation Details}
In Stage A, the probabilistic fluid simulator is trained to predict two future states from two inputs. The ADAM optimizer~\citep{kingma2014adam} is used with an initial learning rate of $0.001$ and a batch size of $16$ for $50k$ iterations.
  We follow previous works~\citep{ummenhofer2020lagrangian, Prantl2022Conserving} to set a scheduled learning rate decay where the learning rate is halved every $5k$ iterations, beginning at iteration $25k$. 
  The latent distribution of each particle is an $8$-dimensional Gaussian with parameterized mean and standard deviation. The KL regularizer $\beta$ is set as $0.1$, shown in Table~\ref{tab:hparams}. The experiments are conducted on 4 NVIDIA RTX 3090 GPUs. 
  To enhance long-term prediction capability, the probabilistic fluid simulator is trained to predict 5 future states from 5 inputs for experiments in visual physical inference. Additionally, we use $\operatorname{tanh}$ as an activation function for layers in the shared feature encoder (CConv). This is to ensure the learned posterior of latent distribution from visual observation lies in the space of the learned physical prior learner.

\rebuttal{\section{Experiments with Non-zero Initial Velocities of the Observed Visual Scene}
To assess the performance of our method in a more general case, specifically, learning from visual observations of fluids with non-zero initial velocities, we modify the training scheme in Stage B by randomly initializing $\{v_{t=1}^i\}_{i=1:N}$ and treating them as trainable parameters. The optimization of $\{v_{t=1}^i\}_{i=1:N}$ is carried out concurrently with the optimization of the visual posteriors $\hat{\mathbf{z}}$. We evaluate the results against a model trained with true non-zero initial velocities given on the observed scene. 
In Table~\ref{tab:init vel obs}, we compare future prediction errors on observed scenes of fluids with non-zero initial velocities. In Table~\ref{tab:init vel novel}, we compare the average simulation errors of the two models on novel scenes with new initial geometries and boundaries. 
We can observe that both models produce comparable results, showcasing the ability of our method to infer uncertain initial velocities of fluids by treating them as optimized variables.}

\begin{table}[ht]
  \centering
  \caption{\rebuttal{Quantitative results of future prediction error $d$ on observed scenes of fluids with non-zero initial velocities. We compare the performance of the model trained with optimized velocity against the model with ground truth non-zero initial velocity given.}}
  \vspace{5pt}
    \begin{small}
  \setlength{\tabcolsep}{1mm}{}
  \begin{sc}
  \resizebox{1.0\linewidth}{!}{
    \begin{tabular}{lcccccc}
    \toprule
    Method & \multicolumn{2}{c}{$\rho=2000,\nu=0.065$} & \multicolumn{2}{c}{$\rho=1000,\nu=0.08$} & \multicolumn{2}{c}{$\rho=500,\nu=0.2$} \\
\cmidrule{2-7}    Trained with true velocity & \multicolumn{2}{c}{\textbf{39.02}} & \multicolumn{2}{c}{\textbf{39.14}} & \multicolumn{2}{c}{\textbf{43.57}} \\
    Trained with optimized velocity & \multicolumn{2}{c}{39.53} & \multicolumn{2}{c}{40.99} & \multicolumn{2}{c}{45.31} \\
    \bottomrule
    \end{tabular}%
    }
    \end{sc}
    \end{small}
  \label{tab:init vel obs}%
\end{table}%

\begin{table}[t]
  \centering
  \caption{\rebuttal{Average prediction error $d$ on novel scenes with new initial geometries and boundaries (denoted by Geom. and Bdy.). Each model is trained with a single visual observation with non-zero velocity on three physical property sets.}}
  \vspace{5pt}
  \begin{small}
  \setlength{\tabcolsep}{1mm}{}
  \begin{sc}
    \begin{tabular}{lcccccc}
    \toprule
          & \multicolumn{2}{c}{$\rho=2000,\nu=0.065$} & \multicolumn{2}{c}{$\rho=1000,\nu=0.08$} & \multicolumn{2}{c}{$\rho=500,\nu=0.2$} \\
\cmidrule{2-7}    Method & \multicolumn{1}{c}{Geom.} & \multicolumn{1}{c}{Bdy.} & \multicolumn{1}{c}{Geom.} & \multicolumn{1}{c}{Bdy.} & \multicolumn{1}{c}{Geom.} & \multicolumn{1}{c}{Bdy.} \\
    Trained with true velocity & 33.39 & 39.74 & \textbf{33.50}  & \textbf{38.41} & 38.52 & 47.23 \\
    Trained with optimized velocity & \textbf{33.25} & \textbf{39.55} & 33.72 & 38.75 & \textbf{38.34} & \textbf{46.90} \\
    \bottomrule
    \end{tabular}%
    \end{sc}
  \end{small}
  \label{tab:init vel novel}%
\end{table}%

\newpage
\section{Additional Visualization Results}
\label{sec:app_add_vis}
\begin{figure*}[t]
\begin{center}
\centerline{
\includegraphics[width=0.99\linewidth]{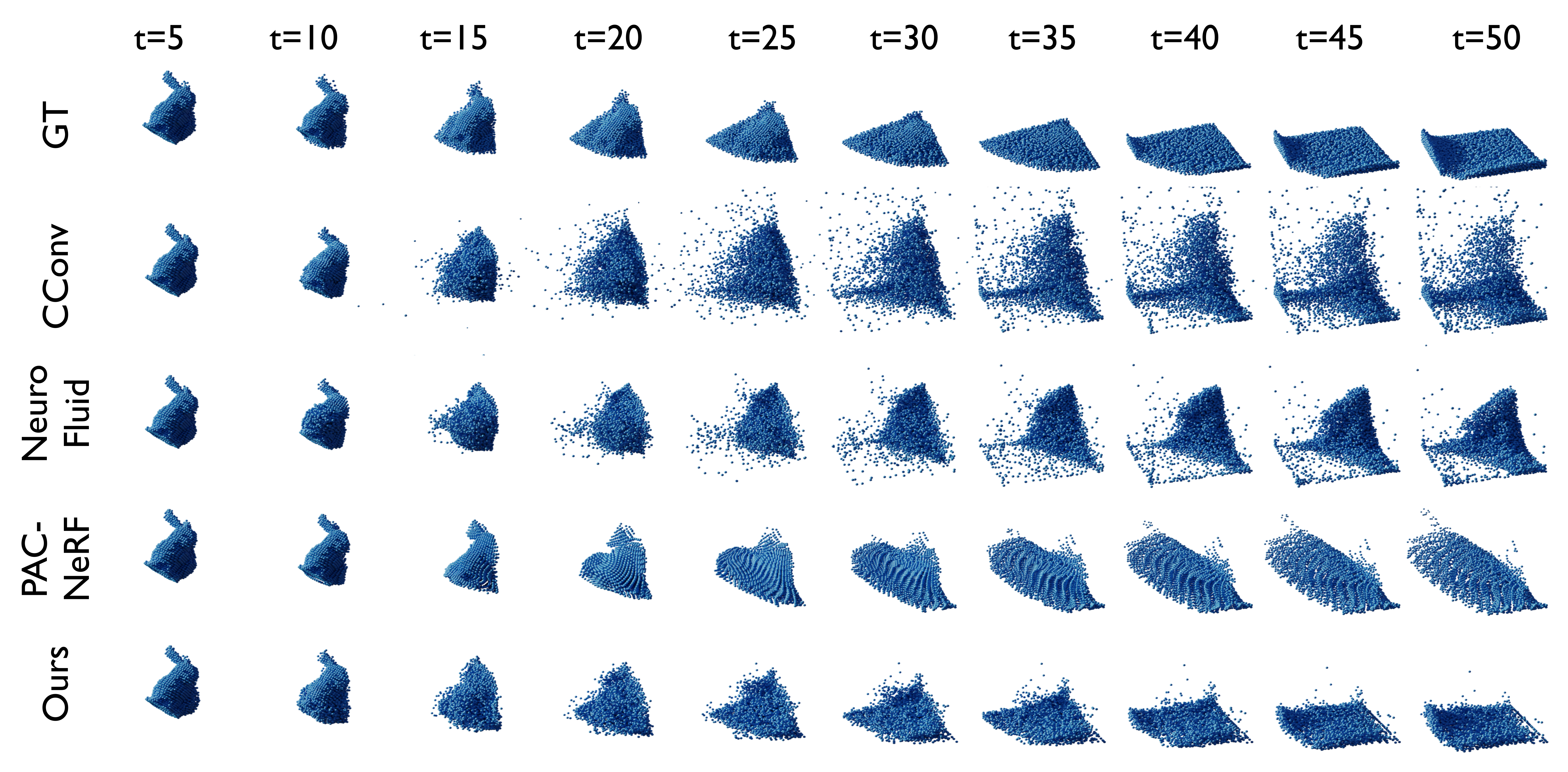}
\vspace{-5pt}
}
\caption{Qualitative results of simulated particles from learned simulators on the unseen fluid geometry (Stanford Bunny) with physical parameter $\rho=2000, \nu=0.065$. In this scenario, the \textit{Stanford Bunny} fluid is randomly sampled with a relatively high horizontal initial velocity and initially makes contact with the side of the container. Due to its low viscosity, the fluid flows rapidly down along the container's side.}
\label{fig: novel2000}
\end{center}
\vskip -0.1in
\end{figure*}

\begin{figure*}[t]
\begin{center}
\centerline{
\includegraphics[width=0.99\linewidth]{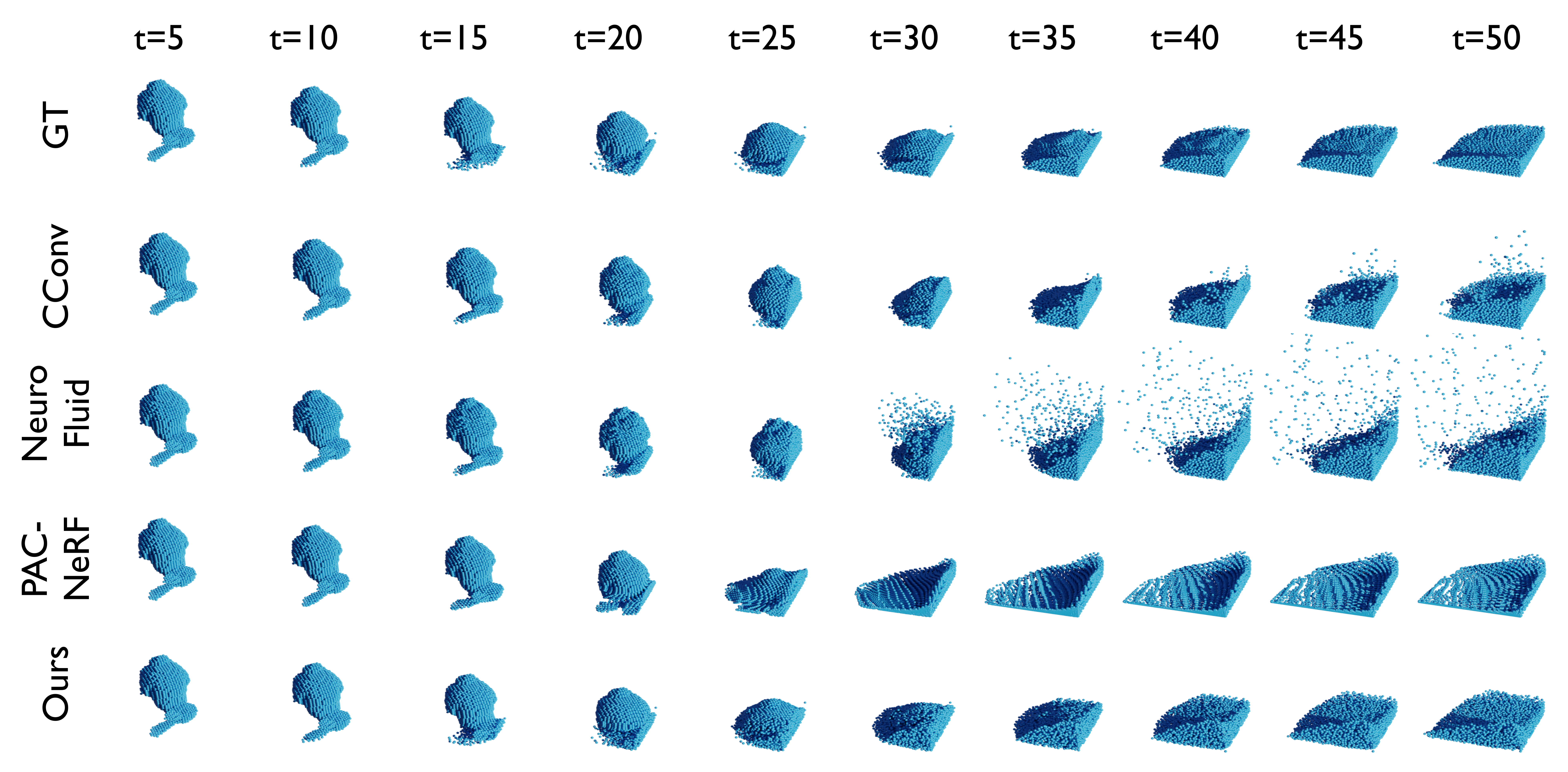}
\vspace{-3pt}
}
\caption{Qualitative results of simulated particles from learned simulators on the unseen fluid geometry (Stanford Bunny) with physical parameter $\rho=1000, \nu=0.08$. In this scenario, the \textit{Stanford Bunny} fluid apply to random rotation, presenting an inverted initial state, and is launched into the container with a relatively low horizontal initial velocity in a projectile motion. }
\label{fig: novel1000}
\end{center}
\vskip -0.1in
\end{figure*}

\begin{figure*}[t]
\begin{center}
\centerline{
\includegraphics[width=0.99\linewidth]{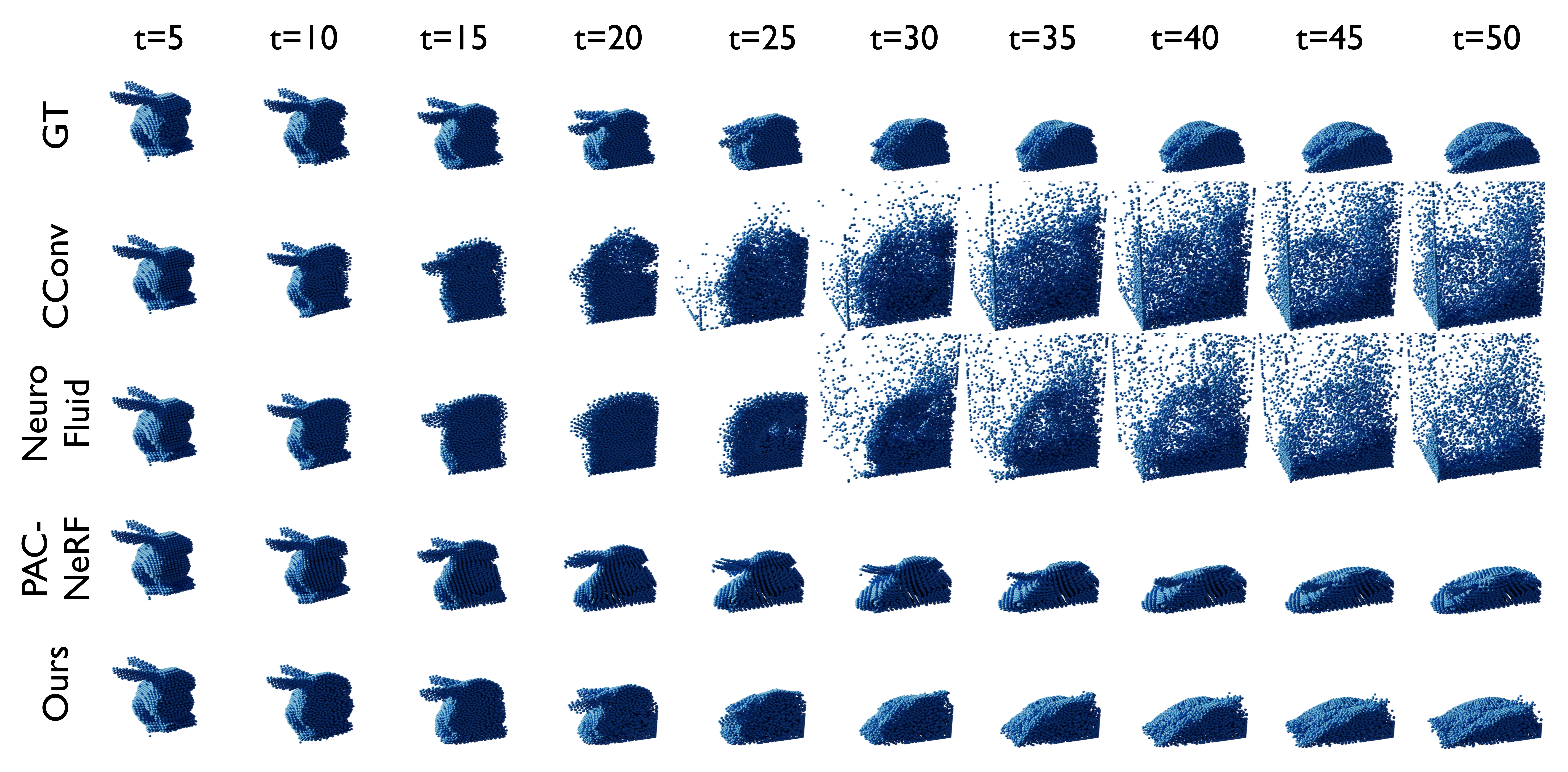}
}
\caption{Qualitative results of simulated particles from learned simulators on the unseen fluid geometry (Stanford Bunny) with physical parameter $\rho=500, \nu=0.2$. In this scenario, the \textit{Stanford Bunny} fluid is randomly sampled with a relatively high horizontal initial velocity and makes initial contact with the side of the container. Due to its high viscosity, the fluid slowly flows down along the container's side. }
\label{fig: novel500}
\end{center}
\vskip -0.1in
\end{figure*}

\begin{figure*}[ht]
\begin{center}
\centerline{
\includegraphics[width=0.99\linewidth]{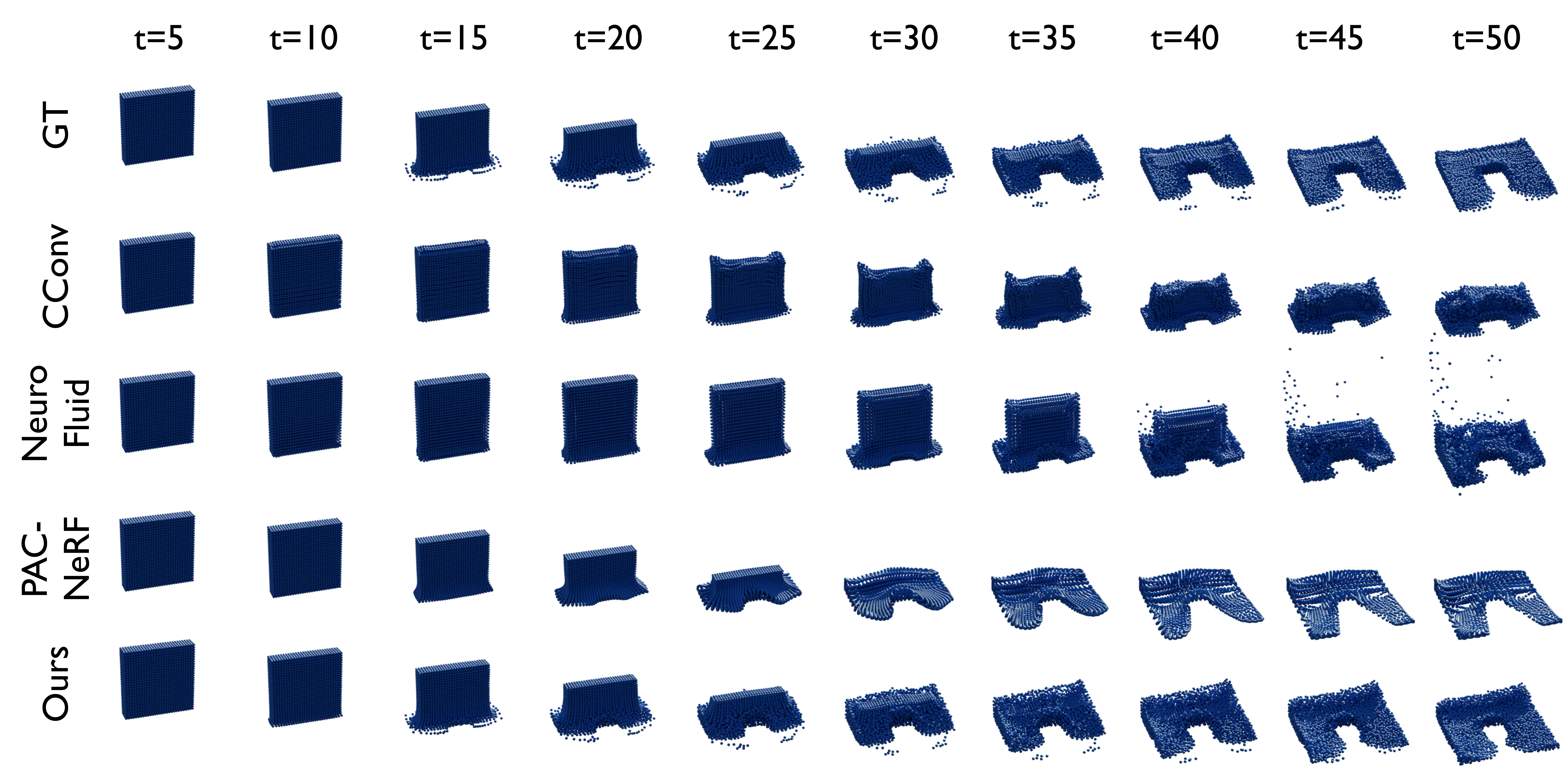}
}
\caption{Qualitative results of simulated particles from learned simulators on the unseen boundary with physical parameter $\rho=2000, \nu=0.065$. In \textit{unseen boundary} scenarios, We use the boundary with a slender pillar positioned in the center. In each scenario, the Dam Break fluid collapses and strikes the pillar in the container, which was unseen during pretraining and from visual observations.}
\label{fig: bnd2000}
\end{center}
\vskip -0.1in
\end{figure*}

\begin{figure*}[t]
\begin{center}
\centerline{
\includegraphics[width=0.99\linewidth]{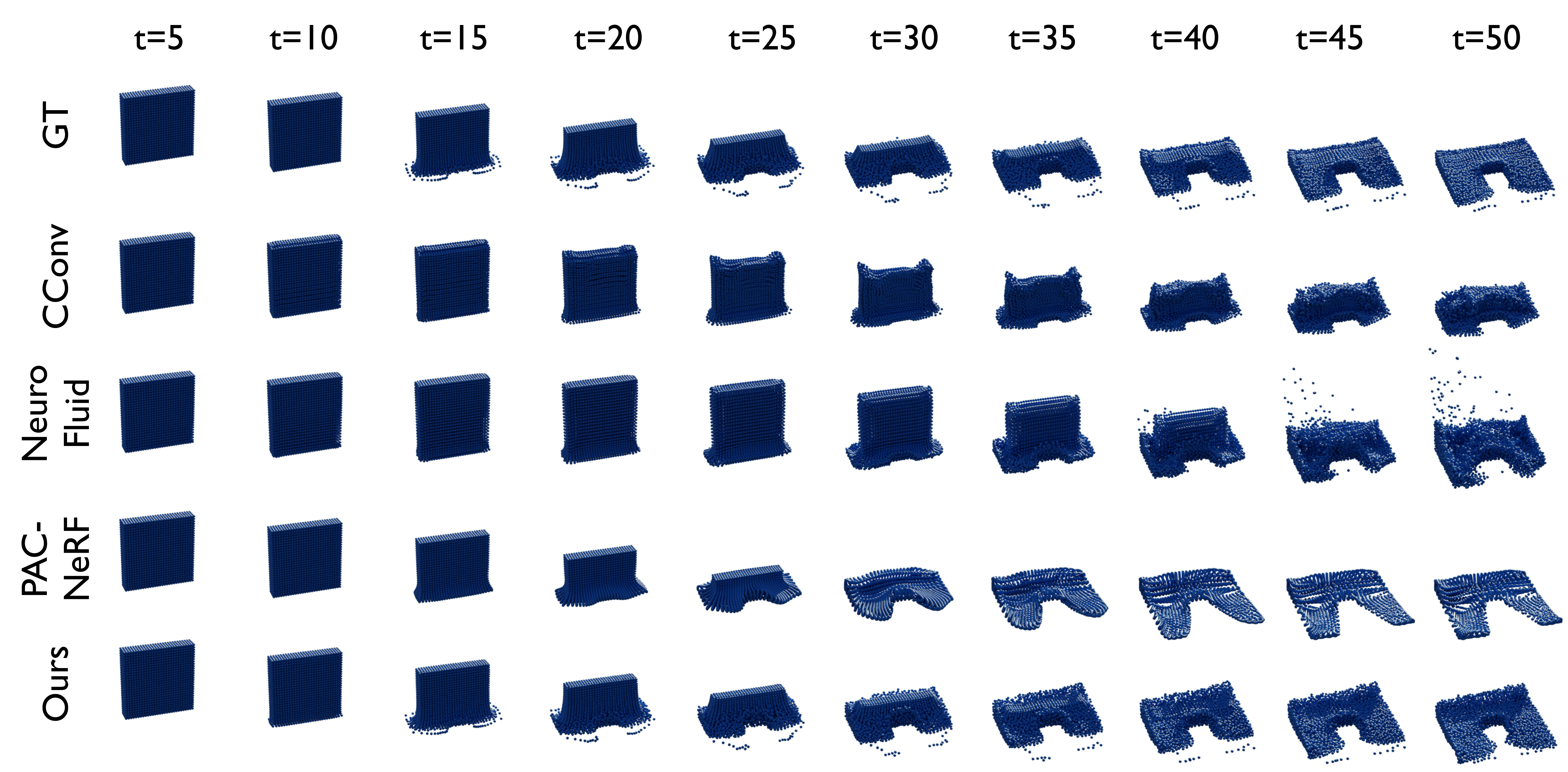}
}
\caption{Qualitative results of simulated particles from learned simulators on the unseen boundary with physical parameter $\rho=1000, \nu=0.08$. PAC-NeRF produces discontinuous simulation predictions.}
\label{fig: bnd1000}
\end{center}
\vskip -0.1in
\end{figure*}

\begin{figure*}[ht]
\begin{center}
\centerline{
\includegraphics[width=0.99\linewidth]{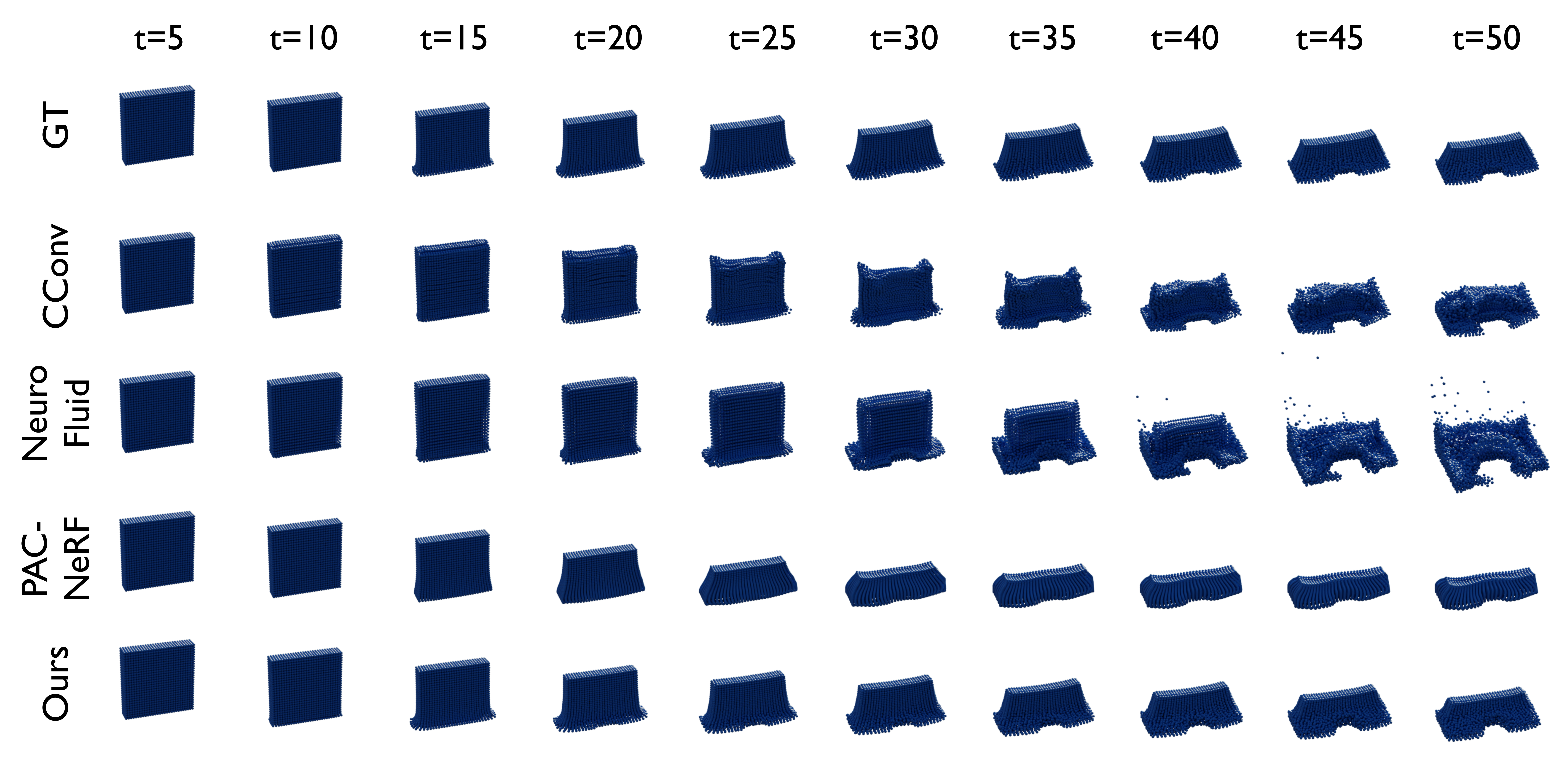}
}
\caption{Qualitative results of simulated particles from learned simulators on the unseen boundary with physical parameter $\rho=500, \nu=0.2$. }
\label{fig: bnd500}
\end{center}
\vskip -0.1in
\end{figure*}

Figure~\ref{fig: novel2000}, \ref{fig: novel1000}, \& \ref{fig: novel500} provide visualizations of the predicted particles of baselines and our model on the novel scenes with unseen fluid geometries. Figure~\ref{fig: bnd2000}, \ref{fig: bnd1000}, \& \ref{fig: bnd500} provide visualizations of the predicted particles of baselines and our model on the novel scenes with unseen boundary conditions. We can see that CConv and NeuroFluid tend to produce noisy predictions on novel scenes. PAC-NeRF produces sporadic predictions that are contrary to the continuous dynamics behavior as ground truth simulations. Our model has more reasonable prediction results and can effectively respond to different physical parameters.

\begin{figure}[t]
\vspace{-6pt}
\begin{center}
\centerline{
\includegraphics[width=\textwidth]{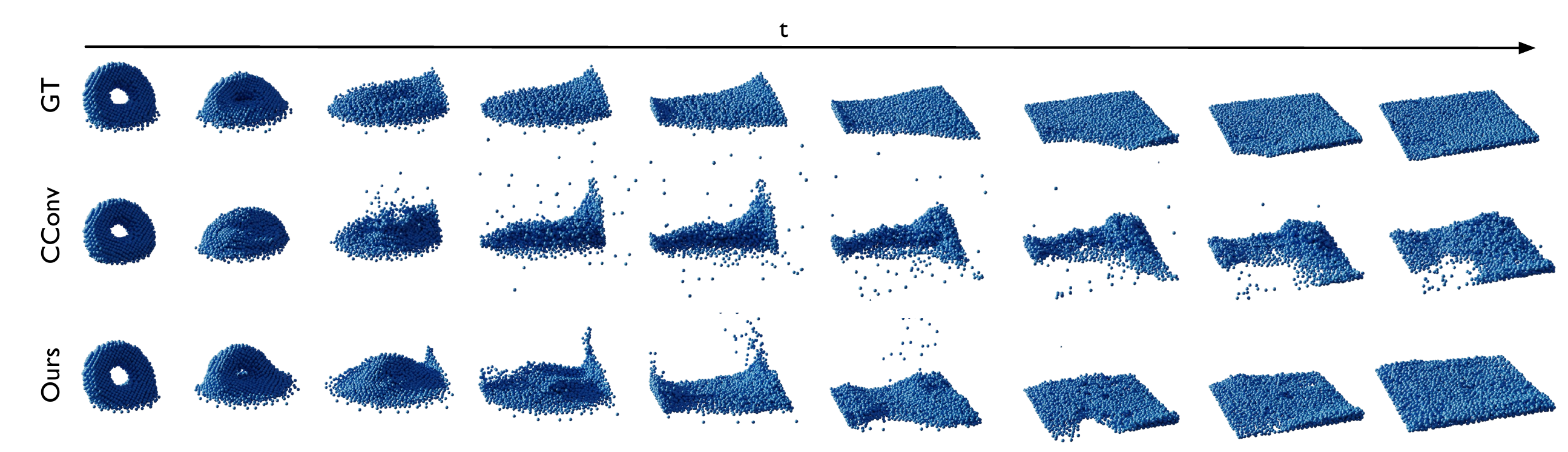}
\vspace{-6pt}
}
\caption{Qualitative results of predicted long-term simulation on the pretrained particle dataset. }
\label{fig: longseq}
\end{center}
\vskip -0.05in
\end{figure}

\clearpage
\newpage
We present qualitative results of the predicted particles of the pretrained probabilistic fluid simulator in Figure~\ref{fig: longseq}. We can see that our method produces closer prediction results with ground truth. This indicates that our probabilistic fluid simulator shows the capability to capture broad hidden physics within the latent space, leveraging spatiotemporal correlation in particle data.

\vspace{-3pt}
\section{Further discussions on Real-world experiments}
\vspace{-3pt}

The primary focus of this paper is to explore the feasibility of the proposed inference--transfer learning scheme for physics, we first use synthetic visual data to evaluate the proposed method. In real-world experiments, however, we need to use specialized fluid flow measurement techniques, such as Particle Image Velocimetry (PIV), to measure the performance of the model.
Most inverse graphics methods are not applied to real-world scenarios of fluid dynamics that undergo intense change and visual noises, including prior arts like NeuroFluid~\citep{guan2022neurofluid} and PAC-NeRF~\citep{li2023pacnerf}. The visual observations of fluid dynamics in real-world scenarios contain more visual noises, such as reflection and refraction, which makes it harder to cope with.

\begin{figure*}[t]
\begin{center}
\vspace{-15pt}
\centerline{
\includegraphics[width=\textwidth]{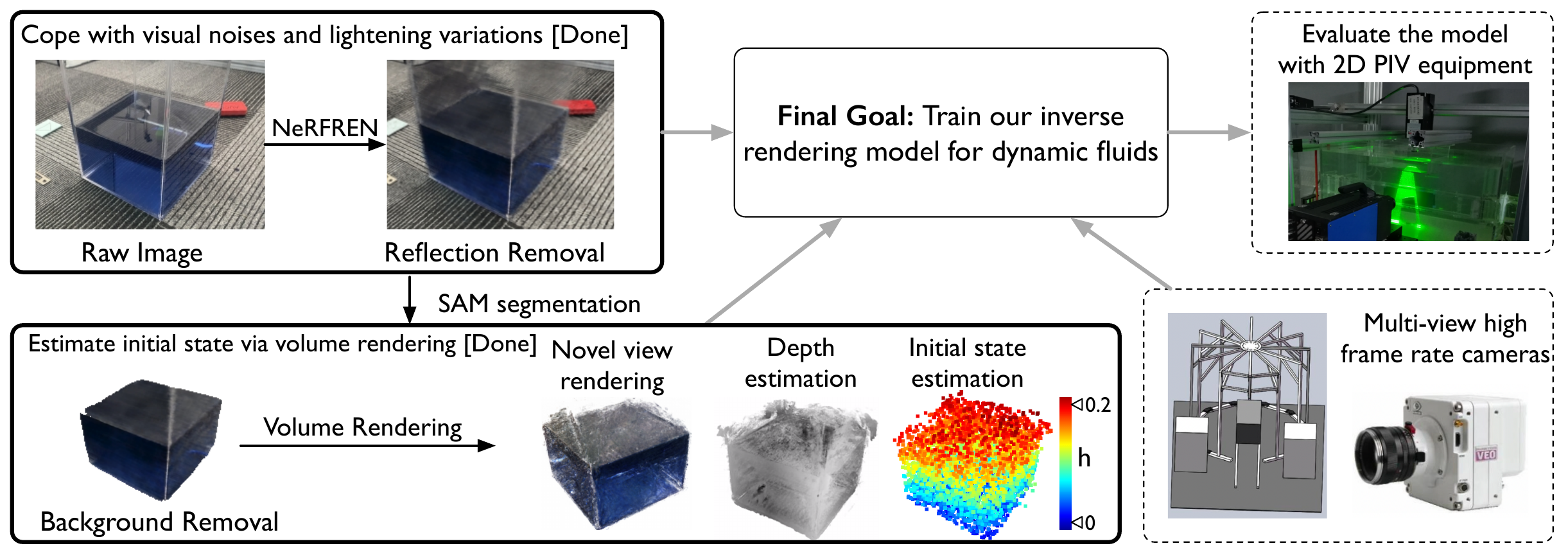}
}
\vspace{-5pt}
\caption{Our pipeline and intermediate results for real-world experiments. Due to the requirements of advanced data acquisition equipment, we leave the complete experiment of dynamic scenes to future work (dashed boxes).}
\label{fig: real world}
\end{center}
\vskip -0.25in
\end{figure*}

Nevertheless, we acknowledge that real-world validation is meaningful and challenging.
To explore the possibility of applications of \textit{\fullname} in real-world scenarios with complex and noisy visual observation, we made our best efforts to conduct real-world experiments. 
As the pipeline provided in Figure~\ref{fig: real world}, we capture the dynamics of dyed water in the fluid tanks. We collect RGB images at a resolution of $1{,}200 \times 900$ on the hemisphere of the scene. 
As the presence of refraction and reflection phenomena (which are usually absent in other dynamics contexts) will bring extra burden for the reconstruction of fluid geometries and dynamics, we first adopt NeRFREN~\citep{Guo_2022_CVPR} to remove the inherent reflection and refraction of fluids and the container. Then we segment the fluid body and remove the background using SAM~\citep{kirillov2023segany}. The preprocessed images are then used to estimate fluid positions using the initial state estimation module. Estimated positions are shown in Figure.~\ref{fig: real world}.
Please refer to the supplementary video for a vivid illustration of real-world experiments. However, an outstanding challenge of the complete experiment is to acquire high frame-rate images with synchronized cameras across multiple viewpoints. Therefore, we have to leave this part for future work.

\label{sec:app_realworld}

\end{document}